\documentclass[%
 reprint,
superscriptaddress,
preprintnumbers,
 amsmath,amssymb,
 aps,
 prl,
]{revtex4-2}

\usepackage{bibunits}
\defaultbibliographystyle{apsrev4-2}
\defaultbibliography{cites}
\usepackage{setspace}

\usepackage{subfiles}
\usepackage{lineno}

\usepackage{graphicx}
\usepackage{dcolumn}
\usepackage{bm}

\renewcommand{\v}[1]{\ensuremath{\mathbf{#1}}} 
 
\renewcommand\eqref[1]{Eq.\;\ref{#1}} 
\newcommand{\equalcontrib}{\thanks{These authors contributed equally to this work.}}

\usepackage [english]{babel}
\usepackage [english = american]{csquotes}
\MakeOuterQuote{"}

\usepackage{color}					

\begin{document}
\begin{bibunit}
\modulolinenumbers[5]

\title{Fractal basins trap latent reasoning}

\author{Jeffrey Lai}\equalcontrib
\affiliation{%
The Oden Institute, The University of Texas at Austin, Austin, Texas 78712, USA
}

\author{Anthony Bao}\equalcontrib
\affiliation{%
Department of Electrical Engineering, The University of Texas at Austin, Austin, Texas 78712, USA
}

\author{John Quinn}
\affiliation{%
The Oden Institute, The University of Texas at Austin, Austin, Texas 78712, USA
}

\author{William Gilpin}
\email{wgilpin@utexas.edu}
\affiliation{%
Department of Physics, The University of Texas at Austin, Austin, Texas 78712, USA
}

\date{\today}

\begin{abstract}
Reasoning allows artificial intelligence models to revisit and correct their mistakes, enabling recent frontier advances in mathematical theorem solving, software engineering, and autonomous task planning.
Reasoning models are widely observed to reason for longer on harder tasks, but the general mechanism responsible for these slowdowns is unknown.
Here, we show that reasoning models exhibit transient chaos, a physical consequence of the computational complexity of difficult tasks. 
As a consequence, we show that diverse leading reasoning models are dynamical systems with fractal basins, with fractality increasing with task difficulty across diverse tasks like Sudoku and maze solving, visual puzzles, and mathematical logic. 
We show that transient chaos emerges due to reasoning becoming trapped for extended durations near saddle points, which we show correspond to nearly-correct attempted solutions of the underlying problem.
Our results show that reasoning slowdowns are an inevitable consequence of problem hardness in modern artificial intelligence models, and establish reasoning traces as a rich new class of dynamical system.
\end{abstract}

\maketitle

\clearpage

Recent advances in frontier artificial intelligence are driven not by scale alone, but rather by the ability to \textit{reason}. Reasoning models revisit and correct earlier mistakes before producing outputs, allowing them to achieve state-of-the-art performance on logical tasks such as visual reasoning, mathematical derivation, and puzzle solving \cite{wei2022chain,guo2025deepseek,snell2025scaling}. 
Reasoning capabilities recently allowed small recurrent models ($7$ million parameters) to outperform large language models (exceeding $10$ billion parameters) on the Abstract Reasoning Corpus (ARC-AGI), a set of complex visual riddles used to benchmark frontier models' fluid intelligence and generalization ability \cite{chollet2019measure,wang2025hierarchicalreasoningmodel,jolicoeurmartineau2025morerecursivereasoningtiny}. 
Yet despite their efficiency, reasoning models are widely reported to suffer from \textit{overthinking}, in which the models become trapped reasoning for extended periods without converging to a solution \cite{chen2025not}. 
Overthinking can nearly double the inference cost of reasoning without improving its accuracy, and adversarially-chosen prompts can induce denial-of-service attacks on frontier reasoning models by consuming $10\times$ more computing resources than nearly-identical benign prompts \cite{liu2026reasoningbomb}.
Overthinking is often attributed informally to the need to explore larger or more uncertain solution spaces \cite{schwarzschild2021can,geiping2025scaling,wang2023selfconsistency}. 
Yet a mechanistic understanding of reasoning dynamics is currently missing, hindering empirical efforts to consistently identify and mitigate reasoning slowdowns \cite{shojaee2026illusion,center2026benchmark,yang2026towards}.

Here, we seek a physical understanding of reasoning using the framework of dynamical systems theory.
Many reasoning models represent tasks as optimization problems: the problem statement and trained model weights parametrize a dynamical system, and the solution represents a stable fixed point \cite{wang2025hierarchicalreasoningmodel,jolicoeurmartineau2025morerecursivereasoningtiny,movahedi2026fixedpointreasonersstableadaptive,huang2026equilibrium}. 
The dynamics either evolve explicitly in the form of intermediate outputs (as in chain-of-thought models) or in the form of a latent encoding of the problem and partial solutions as a hidden state. 
In this sense, modern reasoning models extend efforts to model task-solving using recurrent neural networks, which identify low-dimensional attractors that support computation \cite{sussillo2013opening,khona2022attractor,durstewitz2023reconstructing}.

\begin{figure}[ht]
{
\centering
\includegraphics[width=\linewidth]{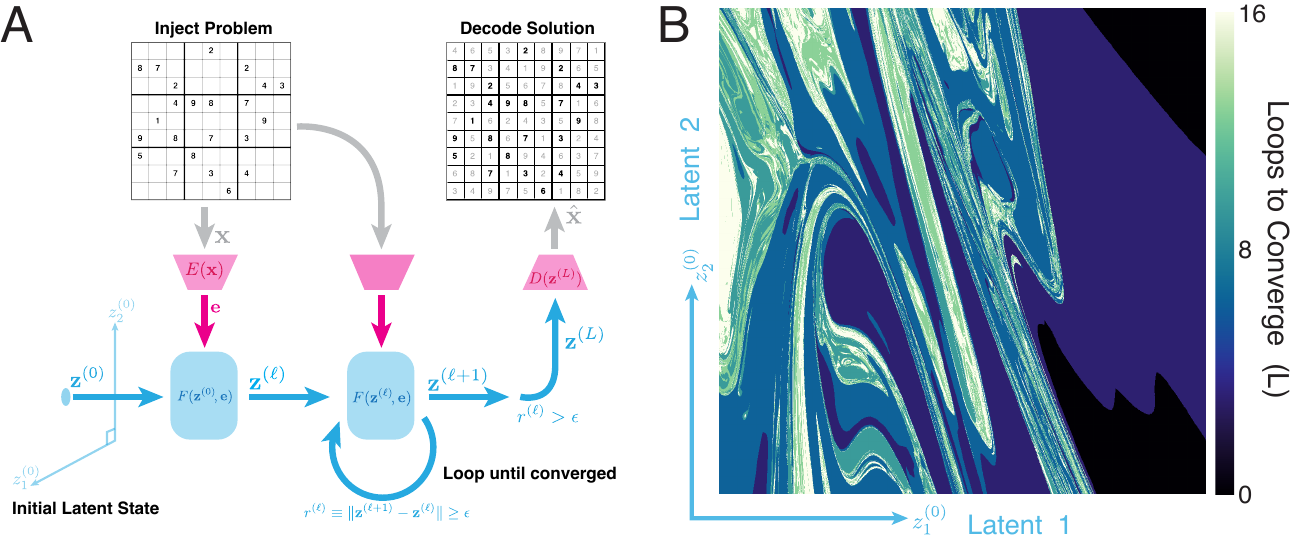}
\caption{
\textbf{Recurrent depth reasoning models produce fractal basins.}
(A) A looped reasoning model encodes a task as a dynamical system evolving in a latent space, and the problem's solution as a fixed point.
(B) Convergence time basins on a difficult Sudoku puzzle not seen during training.
}
\label{fig:basins}
}
\end{figure}

\begin{figure*}[ht]
{
\centering
\includegraphics[width=\linewidth]{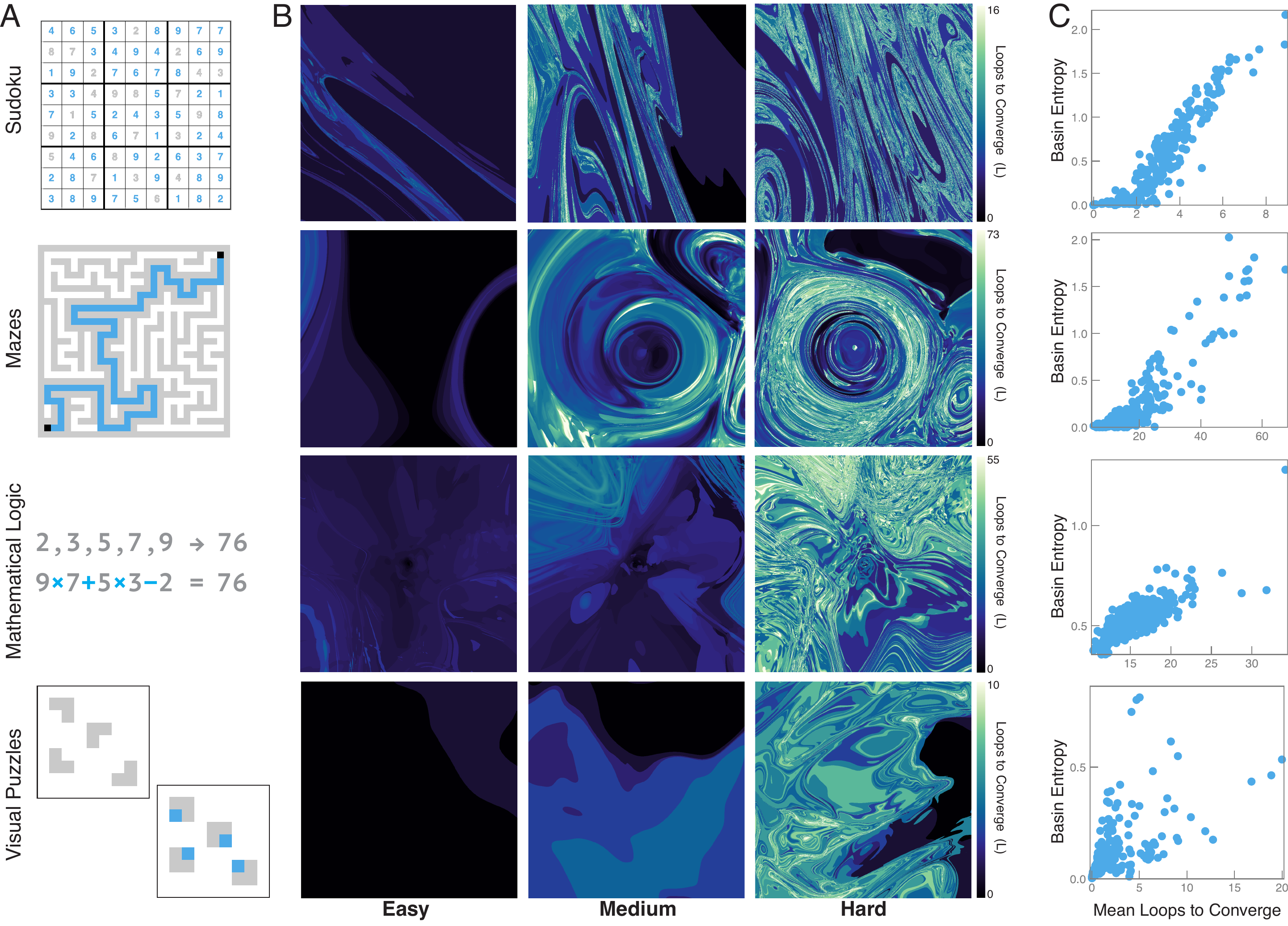}
\caption{
\textbf{Hard problems produce fractal basins across diverse models and tasks.}
(A) Examples of Sudoku-Extreme, Maze-Hard, Countdown, and ARC-AGI-1 reasoning tasks, each of which we probe using a different architecture (Equilibrium Reasoners, Fixed Point Reasoning Models, Parcae, and Tiny Recursive Model) \cite{wang2025hierarchicalreasoningmodel,chollet2019measure,prairie2026parcaescalinglawsstable,jolicoeurmartineau2025morerecursivereasoningtiny,movahedi2026fixedpointreasonersstableadaptive,huang2026equilibrium}
(B) Example basins for varying problem difficulties across the four tasks and architectures.
(C) Scaling of basin entropy with the number of reasoning iterations required to converge, each across randomly-sampled task instances of varying underlying difficulty.
}
\label{fig:diversity}
}
\end{figure*}

The latent state for modern reasoning models is usually initialized arbitrarily, such as by sampling a random vector. We thus introduce a new probe of reasoning dynamics by choosing a model and problem instance, and continuously varying the initial latent state to test how initialization affects convergence across replicates (Fig. \ref{fig:basins}A).
Across diverse reasoning models, we observe complex basins of attraction when we label initial conditions by their convergence time, and we find that these basins become self-similar fractals on harder tasks, like Sudoku puzzles (Fig. \ref{fig:basins}B). 
As a consequence, even an infinitesimal change to the model's initialization increases convergence time by orders of magnitude.
Though reasoning models are trained to solve discrete puzzles, their fractal basins are smooth and naturalistic, resembling optical caustics or transient lifetimes in turbulence---both phenomena arising from spatial variation in the pathlengths required for a physical system to reach fixed points \cite{moehlis2004low,altmann2013leaking}.
In isolated previous cases, fractals have been observed when continuous dynamics are used to represent discrete optimization problems, such as sequence alignment, travelling salesperson routing, and spin glass ground state estimation \cite{hopfield1985neural,chen1995chaotic,elser2007searching}. 
In particular, a previous work derives a continuous-time dynamical system to solve Boolean satisfiability problems (which include Sudoku puzzles) and finds that fractal basins emerge as constraints increase \cite{ercsey2011optimization,ercsey2012chaos}. 
However, here we show that such hardness-induced fractal basins are ubiquitous in modern artificial reasoning models: even in the absence of solvers hand-designed for a particular task, reasoning models produce fractal basins whenever they encounter difficult problems, across highly-distinct model architectures and tasks (Figure \ref{fig:diversity}A).

\begin{figure*}
{
\centering
\includegraphics[width=\linewidth]{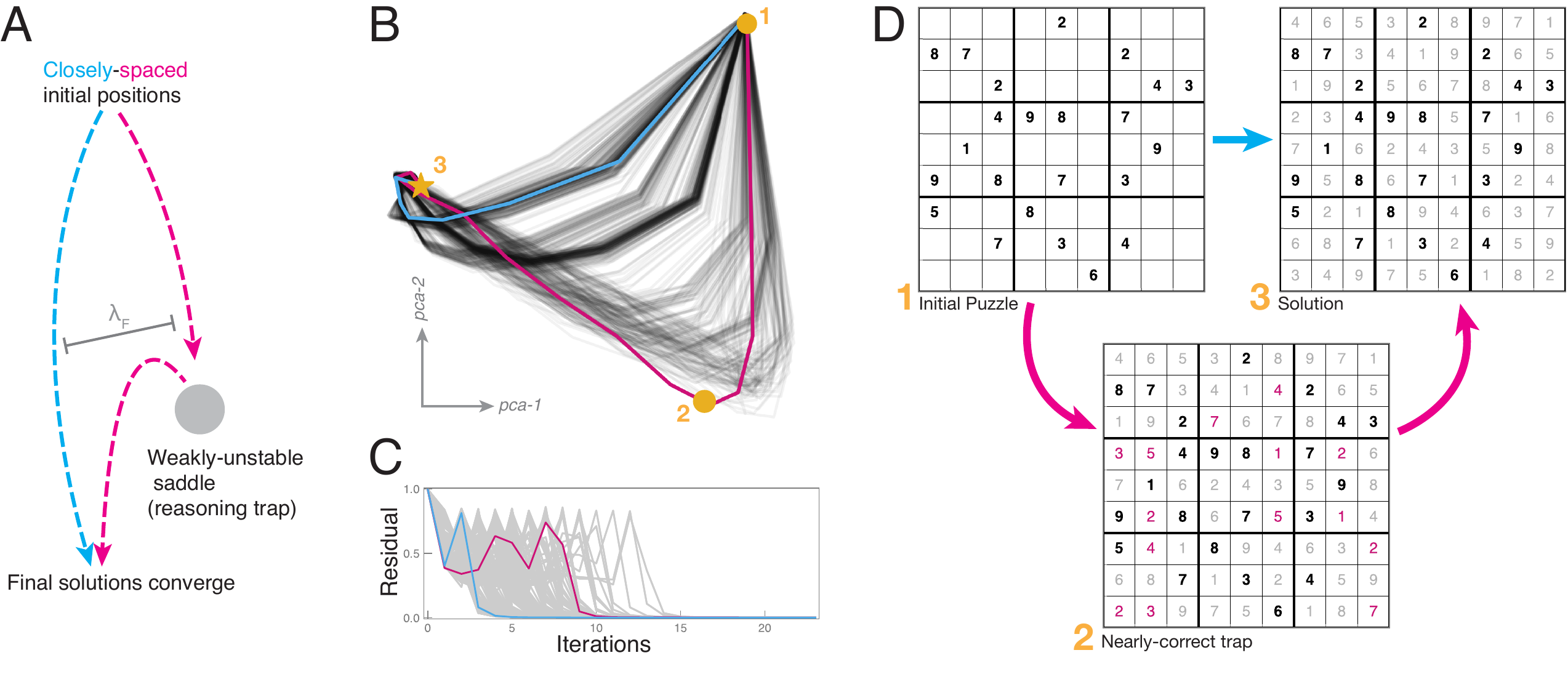}
\caption{
\textbf{Scattering from incorrect solutions produces reasoning transients.} (A) The "Plinko" model of high-dimensional dynamics. Trajectories that start and end at nearly the same point can take very different routes if one encounters weakly-unstable saddle points. The fast Lyapunov indicator ($\lambda_F$) measures their maximum separation. 
(B) The latent state dynamics projected into two dimensions using PCA, showing two representative initial conditions that start out closely-spaced, but which take direct and indirect routes scattered by a saddle point.
(C) The error between the latent state and converged latent state for trajectories originating from the initial conditions in (B). 
(D) Decoding the latent state near the saddle point produces a nearly-correct solution attempt.
}
\label{fig:scatter}
}
\end{figure*}

We determine the prevalence of fractal basins using basin entropy, a recently-introduced metric that distinguishes self-similar fractal basins from random or smooth basins \cite{daza2016basin}. 
For each reasoning model and task, we randomly sample initial latent states along randomly-oriented two-dimensional slices through the full initial latent space. For each slice, we measure the average number of loops required for the model to converge across all initial conditions, and compare it to the basin entropy of the entire slice (Fig. \ref{fig:diversity}C). 
We find that the basin entropy strongly correlates with the number of iterations, a measure of overall task difficulty.
These findings persist across alternative architectures (\ref{app:replicates}) and basin complexity metrics (\ref{app:metrics}).

Fractal basins form due to transient chaos, the phenomenon responsible for extended transients in high-dimensional dynamical systems \cite{grebogi1983fractal,tel2008chaotic,motter2013doubly}. 
Successfully solving a hard problem requires that all trajectories of a reasoning model eventually converge to a single fixed point---and thus that no sustained chaos occurs. 
However, trajectories can take widely-varying routes to equilibrium, due to the presence of saddle points (weakly-unstable solutions) scattered throughout phase space, that redirect trajectories for extended durations and thus delay their convergence to equilibrium. 
High-dimensional dynamics thus resemble a "Plinko" toy: initially closely-spaced initial conditions scatter off of weakly-unstable saddle sets, which pull trajectories apart and create diverging routes to equilibrium (Fig. \ref{fig:scatter}A) \cite{dauphin2014identifying}. 
The scrambling of trajectories caused by saddles produces the complex structure of basins.

\begin{figure}
{
\centering
\includegraphics[width=\linewidth]{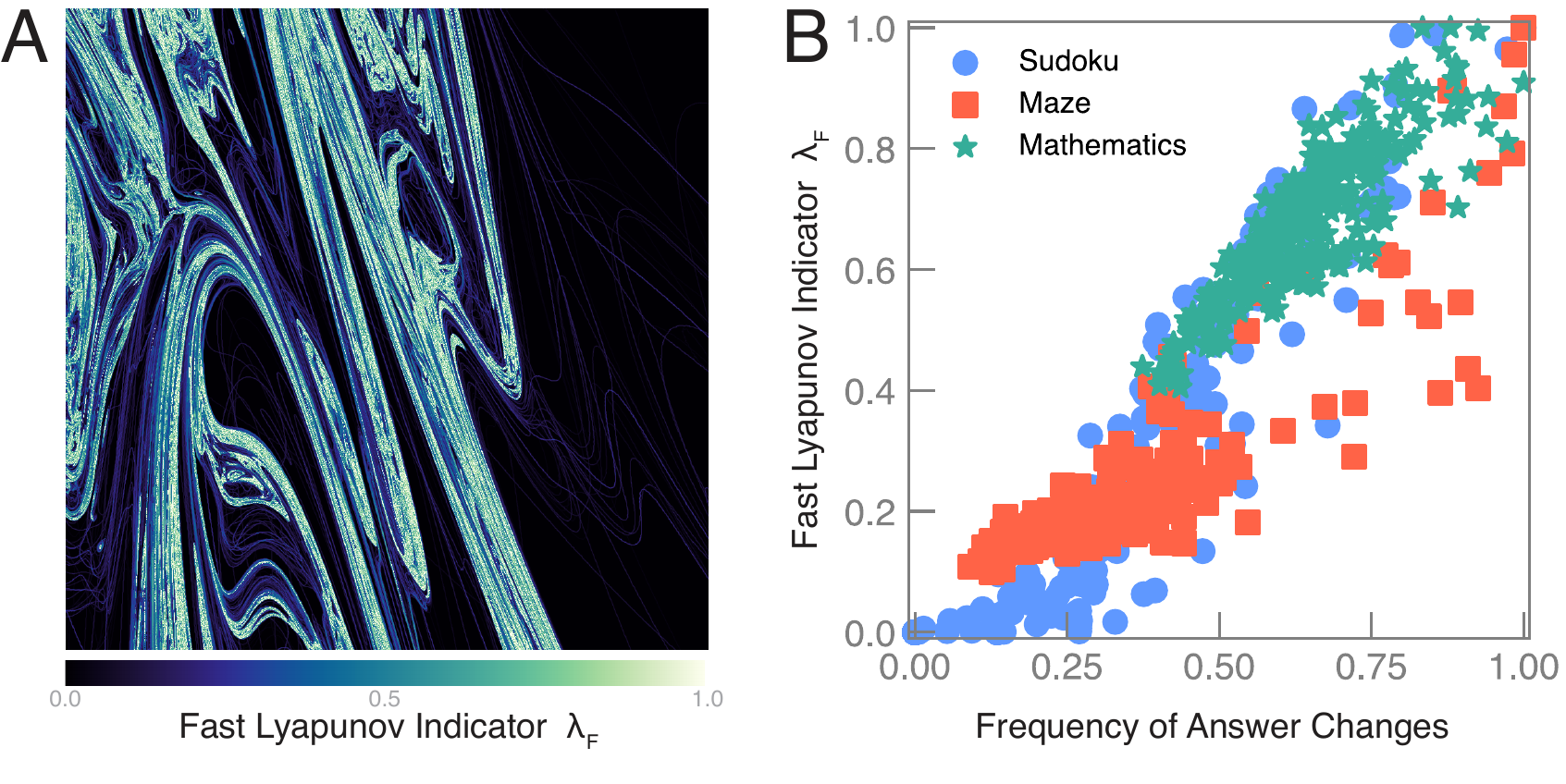}
\caption{
\textbf{Saddle sets indicate uncertain reasoning trajectories.}
(A) The fast Lyapunov Indicator $\lambda_F$ versus initial condition for a Sudoku-solving model, with two representative initial conditions highlighted (magenta and blue crosses). 
(B) $\lambda_F$ versus the frequency at which reasoning traces switch solutions. We measure the frequency by decoding the latent reasoning dynamics to a candidate solution at each timestep, and measuring the number of changes divided by the total number of iterations before convergence. Each point corresponds to a different puzzle instance.
}
\label{fig:fli}
}
\end{figure}

We thus consider the underlying cause of saddle formation in reasoning tasks.
We project the sequence of latent states encountered during high-dimensional reasoning traces into fewer dimensions using principal components analysis. 
We find that faster-converging trajectories take direct paths toward the true solution, while slower trajectories take longer, indirect routes that visit saddle points before reaching the true solution (Fig. \ref{fig:scatter}B).
Decoding these meandering trajectories into problem states reveals that reasoning becomes trapped near answers that are nearly, but not quite, correct. 
In mazes, saddle regions correspond to dead ends; in Sudoku puzzles, they represent grids with repeated digits (Fig. \ref{fig:scatter}D).
The amount of time that a dynamical system spends near a saddle depends on the number of outgoing escape directions \cite{kantz1985repellers}. In gradient systems, this corresponds to the number of downhill directions.

\begin{figure*}[ht]
{
\centering
\includegraphics[width=\linewidth]{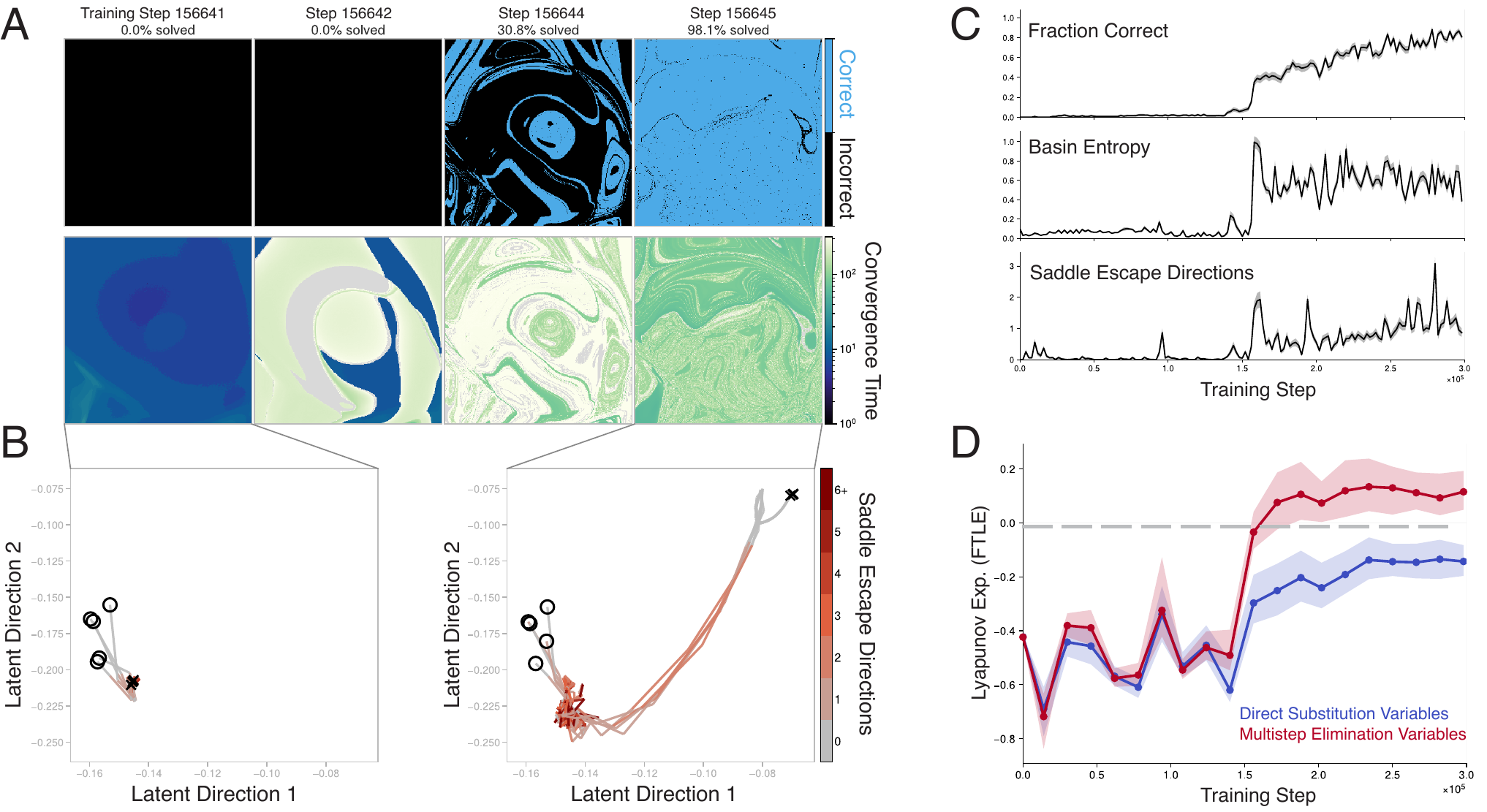}
\caption{
\textbf{Fractal basins emerge as bifurcations during training.}
(A) A looped transformer is trained to solve integer linear systems. Panels correspond to a series of training steps around the bifurcation that occurs $150$k steps into training, when accurate solving first emerges. The top row colors initial conditions based on whether they converge to the correct solution, and the bottom row colors trajectories by their convergence time.
(B) Latent reasoning trajectories before and after the bifurcation, projected into two latent components. Open circles denote starting points, and crosses denote final fixed points.
(C) The fraction of correct solutions, basin entropy, and number of saddle escape directions along trajectories across the full training procedure.
(D) The finite-time Lyapunov exponents separately-calculated for the directly-substitutable variables, and more difficult core variables that require multistep Gaussian elimination to solve.
}
\label{fig:mechanism}
}
\end{figure*}

We directly measure the saddle sets responsible for transient chaos using the fast Lyapunov indicator $\lambda_F$, a quantity originally developed to measure the stability of asteroid orbits \cite{froeschle1997fast}. 
Two initially-close trajectories that take different routes to equilibrium have larger $\lambda_F$, implying that they straddle opposite sides of at least one saddle. 
On single puzzles, $\lambda_F$ reveals boundaries between different solution routes in reasoning traces (Fig. \ref{fig:fli}A). From an algorithmic perspective, these represent maximally-uncertain solution routes. Across many different puzzles for the three different models considered here, we find that the fast Lyapunov indicator strongly correlates with the number of distinct solutions that a given reasoning trajectory passes through before converging. Thus, saddle points shape the reasoning landscape by encoding the solution structure of the underlying problem (Fig. \ref{fig:fli}B).

Generally, when high-dimensional dynamical systems transition from a complex phase to a monostable phase, their local minima disappear, leaving a single stable equilibrium but many weakly-unstable saddle points \cite{ben2021counting,wainrib2013topological}.
In reasoning tasks, this transition occurs when an underconstrained puzzle with multiple solutions (such as a maze with two routes) loses a solution due to the addition of a new constraint (such as an additional wall blocking an exit route).
The former solution still affects the convergence time of solvers, which may traverse all the way through the previous solution before reaching the barrier, requiring backtracking \cite{dechter1994experimental}.
Generally, constraint-solving problems are often hardest at this constrainedness transition: the problem is still solvable, but with few distinct routes to the correct solution---thus increasing the amount of required trial-and-error \cite{cheeseman1991really,monasson1999determining}.
We find that saddles encode these dead end reasoning routes in reasoning dynamical systems.

To study the emergence of saddle-mediated fractal basins, we train a miniaturized reasoning model to solve integer linear systems: For prime $p$, given $\v{A} \in \mathbb{F}_p^{M\times N}, \v{b}\in\mathbb{F}_p^{M}$, solve $\v{A} \v{x} = \v{b}$ for $\v{x}\in\mathbb{F}_p^N$. There are $p^N$ candidate solutions corresponding to all possible $N$-dimensional integer-valued vectors over $\mathbb{F}_p = \{0, 1, \ldots, p-1\}$. 
Typical discrete algorithms for this problem test sequences of integers until a constraint is found violated. 
We train an encoder-only looped transformer on many random instances with $p=3, M=N=8$, and at each stage of training we analyze the reasoning dynamics on newly sampled problem instances.
In early stages of training, the model exhibits a basin landscape resembling that of a random dynamical system \cite{wainrib2013topological}, with basins surrounding distinct stable fixed points, most of which represent incorrect solutions (Fig. \ref{fig:mechanism}A). 
However, in this regime, the convergence times do not show fractal basins. 
As training proceeds, a bifurcation occurs, with the model's weights acting as a control parameter: the incorrect solutions lose stability and become saddles, and reasoning trajectories begin converging to the fixed point of the true solution (Fig. \ref{fig:mechanism}A). 
In projections of the latent space, trajectories transition from smoothly converging to their nearest (incorrect) solution to taking more circuitous routes that pass near saddle points (Fig. \ref{fig:mechanism}B).
As a result, at the bifurcation we observe an abrupt increase in solvability and basin entropy, and the appearance of unstable saddle directions, as measured by the number of unstable eigenvalues in the Jacobian of the reasoning model (Fig. \ref{fig:mechanism}C).
These unstable directions confirm that local minima associated with incorrect solutions lose stability. However, saddles still weakly attract trajectories, producing basins in the convergence time field that resemble the former basins of incorrect solutions.
Such saddle-mediated bifurcations produce fractal basins in many classical nonlinear systems \cite{grebogi1983fractal}. 
We calculate the spectrum of finite-time Lyapunov exponents (FTLE), a set of rates that quantify how small perturbations to the latent state are initially amplified or suppressed along independent directions \cite{lai2011transient}. Pre-bifurcation, all FTLE are negative and so no transient chaos occurs: trajectories smoothly approach nearby local minima representing incorrect solutions. After the bifurcation, unstable directions appear, and transient chaos emerges due to scattering from saddle points.
To interpret the unstable directions, we partition the $N$ variables of the original problem into a subset of simple variables solvable with direct substitution, and core variables which require multi-step Gaussian elimination to resolve. 
This partitioning is analogous to constraint satisfaction problems with leaf variables (resolvable by constraint propagation) and the core (the sub-problem remaining after constraint propagation) \cite{braunstein2002complexity}. 
Only the core variables produce positive FTLE, implying that the algorithm that the model uses to resolve these variables produces transient chaos (Fig. \ref{fig:mechanism}D). We thus find that the bifurcation to solvability during training occurs when the model gains multi-step reasoning capability. This, in turn, allows the model to escape from incorrect solutions in the core, but it leads to transient chaos and fractal basins.


We have discovered that diverse reasoning models navigate fractal basins when they solve hard problems. 
Our observations hold across diverse model architectures and tasks, underscoring that fractality is a general phenomenon in learning models, arising from the algorithms reasoning models learn to apply during inference.
Practically, our findings introduce a new probe of hidden reasoning processes, and show that models' capabilities are related to their ability to escape saddle points near incorrect solutions.
Our findings support a general interpretation of fractal basins and transient chaos as the generic manner in which computational complexity manifests in analogue systems \cite{moore1990unpredictability}, thus connecting the classical physics of nonlinear dynamics to the discrete symbolic reasoning tasks encountered by modern artificial intelligence models. 
Our results thus place modern reasoning models within a longstanding tradition framing computation as a physical process \cite{shaw1981strange,crutchfield1989inferring}---supporting a view of complexity as a physical quantity, with consequences for inference as artificial intelligence methods continue to evolve.

\section{Acknowledgments}

We thank Unconventional AI for supporting our group. 
This work was performed in part at the Aspen Center for Physics, which is supported by NSF grant PHY-2210452.
W.G. is supported by NSF DMS 2436233 and NSF CMMI 2440490, Grant No. DAF2023-329596 from the Chan Zuckerberg Initiative DAF (an advised fund of Silicon Valley Community Foundation), and grant CS-CSA-2026-075 from Research Corporation for Science Advancement. 
Computational analyses were performed using the Biomedical Research Computing Facility at UT Austin, Center for Biomedical Research Support (RRID: SCR\_021979) and the Texas Advanced Computing Center (TACC).


\begin{thebibliography}{44}%
\makeatletter
\providecommand \@ifxundefined [1]{%
 \@ifx{#1\undefined}
}%
\providecommand \@ifnum [1]{%
 \ifnum #1\expandafter \@firstoftwo
 \else \expandafter \@secondoftwo
 \fi
}%
\providecommand \@ifx [1]{%
 \ifx #1\expandafter \@firstoftwo
 \else \expandafter \@secondoftwo
 \fi
}%
\providecommand \natexlab [1]{#1}%
\providecommand \enquote  [1]{``#1''}%
\providecommand \bibnamefont  [1]{#1}%
\providecommand \bibfnamefont [1]{#1}%
\providecommand \citenamefont [1]{#1}%
\providecommand \href@noop [0]{\@secondoftwo}%
\providecommand \href [0]{\begingroup \@sanitize@url \@href}%
\providecommand \@href[1]{\@@startlink{#1}\@@href}%
\providecommand \@@href[1]{\endgroup#1\@@endlink}%
\providecommand \@sanitize@url [0]{\catcode `\\12\catcode `\$12\catcode `\&12\catcode `\#12\catcode `\^12\catcode `\_12\catcode `\%12\relax}%
\providecommand \@@startlink[1]{}%
\providecommand \@@endlink[0]{}%
\providecommand \url  [0]{\begingroup\@sanitize@url \@url }%
\providecommand \@url [1]{\endgroup\@href {#1}{\urlprefix }}%
\providecommand \urlprefix  [0]{URL }%
\providecommand \Eprint [0]{\href }%
\providecommand \doibase [0]{https://doi.org/}%
\providecommand \selectlanguage [0]{\@gobble}%
\providecommand \bibinfo  [0]{\@secondoftwo}%
\providecommand \bibfield  [0]{\@secondoftwo}%
\providecommand \translation [1]{[#1]}%
\providecommand \BibitemOpen [0]{}%
\providecommand \bibitemStop [0]{}%
\providecommand \bibitemNoStop [0]{.\EOS\space}%
\providecommand \EOS [0]{\spacefactor3000\relax}%
\providecommand \BibitemShut  [1]{\csname bibitem#1\endcsname}%
\let\auto@bib@innerbib\@empty
\bibitem [{\citenamefont {Wei}\ \emph {et~al.}(2022)\citenamefont {Wei}, \citenamefont {Wang}, \citenamefont {Schuurmans}, \citenamefont {Bosma}, \citenamefont {Xia}, \citenamefont {Chi}, \citenamefont {Le}, \citenamefont {Zhou} \emph {et~al.}}]{wei2022chain}%
  \BibitemOpen
  \bibfield  {author} {\bibinfo {author} {\bibfnamefont {J.}~\bibnamefont {Wei}}, \bibinfo {author} {\bibfnamefont {X.}~\bibnamefont {Wang}}, \bibinfo {author} {\bibfnamefont {D.}~\bibnamefont {Schuurmans}}, \bibinfo {author} {\bibfnamefont {M.}~\bibnamefont {Bosma}}, \bibinfo {author} {\bibfnamefont {F.}~\bibnamefont {Xia}}, \bibinfo {author} {\bibfnamefont {E.}~\bibnamefont {Chi}}, \bibinfo {author} {\bibfnamefont {Q.~V.}\ \bibnamefont {Le}}, \bibinfo {author} {\bibfnamefont {D.}~\bibnamefont {Zhou}}, \emph {et~al.},\ }\href@noop {} {\bibfield  {journal} {\bibinfo  {journal} {Advances in neural information processing systems}\ }\textbf {\bibinfo {volume} {35}},\ \bibinfo {pages} {24824} (\bibinfo {year} {2022})}\BibitemShut {NoStop}%
\bibitem [{\citenamefont {Guo}\ \emph {et~al.}(2025)\citenamefont {Guo}, \citenamefont {Yang}, \citenamefont {Zhang}, \citenamefont {Song}, \citenamefont {Wang}, \citenamefont {Zhu}, \citenamefont {Xu}, \citenamefont {Zhang}, \citenamefont {Ma}, \citenamefont {Bi} \emph {et~al.}}]{guo2025deepseek}%
  \BibitemOpen
  \bibfield  {author} {\bibinfo {author} {\bibfnamefont {D.}~\bibnamefont {Guo}}, \bibinfo {author} {\bibfnamefont {D.}~\bibnamefont {Yang}}, \bibinfo {author} {\bibfnamefont {H.}~\bibnamefont {Zhang}}, \bibinfo {author} {\bibfnamefont {J.}~\bibnamefont {Song}}, \bibinfo {author} {\bibfnamefont {P.}~\bibnamefont {Wang}}, \bibinfo {author} {\bibfnamefont {Q.}~\bibnamefont {Zhu}}, \bibinfo {author} {\bibfnamefont {R.}~\bibnamefont {Xu}}, \bibinfo {author} {\bibfnamefont {R.}~\bibnamefont {Zhang}}, \bibinfo {author} {\bibfnamefont {S.}~\bibnamefont {Ma}}, \bibinfo {author} {\bibfnamefont {X.}~\bibnamefont {Bi}}, \emph {et~al.},\ }\href@noop {} {\bibfield  {journal} {\bibinfo  {journal} {Nature}\ }\textbf {\bibinfo {volume} {645}},\ \bibinfo {pages} {633} (\bibinfo {year} {2025})}\BibitemShut {NoStop}%
\bibitem [{\citenamefont {Snell}\ \emph {et~al.}(2025)\citenamefont {Snell}, \citenamefont {Lee}, \citenamefont {Xu},\ and\ \citenamefont {Kumar}}]{snell2025scaling}%
  \BibitemOpen
  \bibfield  {author} {\bibinfo {author} {\bibfnamefont {C.~V.}\ \bibnamefont {Snell}}, \bibinfo {author} {\bibfnamefont {J.}~\bibnamefont {Lee}}, \bibinfo {author} {\bibfnamefont {K.}~\bibnamefont {Xu}},\ and\ \bibinfo {author} {\bibfnamefont {A.}~\bibnamefont {Kumar}},\ }in\ \href {https://openreview.net/forum?id=4FWAwZtd2n} {\emph {\bibinfo {booktitle} {The Thirteenth International Conference on Learning Representations}}}\ (\bibinfo {year} {2025})\BibitemShut {NoStop}%
\bibitem [{\citenamefont {Chollet}(2019)}]{chollet2019measure}%
  \BibitemOpen
  \bibfield  {author} {\bibinfo {author} {\bibfnamefont {F.}~\bibnamefont {Chollet}},\ }\href@noop {} {\bibfield  {journal} {\bibinfo  {journal} {arXiv preprint arXiv:1911.01547}\ } (\bibinfo {year} {2019})}\BibitemShut {NoStop}%
\bibitem [{\citenamefont {Wang}\ \emph {et~al.}(2025)\citenamefont {Wang}, \citenamefont {Li}, \citenamefont {Sun}, \citenamefont {Chen}, \citenamefont {Liu}, \citenamefont {Wu}, \citenamefont {Lu}, \citenamefont {Song},\ and\ \citenamefont {Yadkori}}]{wang2025hierarchicalreasoningmodel}%
  \BibitemOpen
  \bibfield  {author} {\bibinfo {author} {\bibfnamefont {G.}~\bibnamefont {Wang}}, \bibinfo {author} {\bibfnamefont {J.}~\bibnamefont {Li}}, \bibinfo {author} {\bibfnamefont {Y.}~\bibnamefont {Sun}}, \bibinfo {author} {\bibfnamefont {X.}~\bibnamefont {Chen}}, \bibinfo {author} {\bibfnamefont {C.}~\bibnamefont {Liu}}, \bibinfo {author} {\bibfnamefont {Y.}~\bibnamefont {Wu}}, \bibinfo {author} {\bibfnamefont {M.}~\bibnamefont {Lu}}, \bibinfo {author} {\bibfnamefont {S.}~\bibnamefont {Song}},\ and\ \bibinfo {author} {\bibfnamefont {Y.~A.}\ \bibnamefont {Yadkori}},\ }\href {https://arxiv.org/abs/2506.21734} {\bibinfo {title} {Hierarchical reasoning model}} (\bibinfo {year} {2025}),\ \Eprint {https://arxiv.org/abs/2506.21734} {arXiv:2506.21734 [cs.AI]} \BibitemShut {NoStop}%
\bibitem [{\citenamefont {Jolicoeur-Martineau}(2025)}]{jolicoeurmartineau2025morerecursivereasoningtiny}%
  \BibitemOpen
  \bibfield  {author} {\bibinfo {author} {\bibfnamefont {A.}~\bibnamefont {Jolicoeur-Martineau}},\ }\href {https://arxiv.org/abs/2510.04871} {\bibinfo {title} {Less is more: Recursive reasoning with tiny networks}} (\bibinfo {year} {2025}),\ \Eprint {https://arxiv.org/abs/2510.04871} {arXiv:2510.04871 [cs.LG]} \BibitemShut {NoStop}%
\bibitem [{\citenamefont {Chen}\ \emph {et~al.}(2025)\citenamefont {Chen}, \citenamefont {Xu}, \citenamefont {Liang}, \citenamefont {He}, \citenamefont {Pang}, \citenamefont {Yu}, \citenamefont {Song}, \citenamefont {Liu}, \citenamefont {Zhou}, \citenamefont {Zhang} \emph {et~al.}}]{chen2025not}%
  \BibitemOpen
  \bibfield  {author} {\bibinfo {author} {\bibfnamefont {X.}~\bibnamefont {Chen}}, \bibinfo {author} {\bibfnamefont {J.}~\bibnamefont {Xu}}, \bibinfo {author} {\bibfnamefont {T.}~\bibnamefont {Liang}}, \bibinfo {author} {\bibfnamefont {Z.}~\bibnamefont {He}}, \bibinfo {author} {\bibfnamefont {J.}~\bibnamefont {Pang}}, \bibinfo {author} {\bibfnamefont {D.}~\bibnamefont {Yu}}, \bibinfo {author} {\bibfnamefont {L.}~\bibnamefont {Song}}, \bibinfo {author} {\bibfnamefont {Q.}~\bibnamefont {Liu}}, \bibinfo {author} {\bibfnamefont {M.}~\bibnamefont {Zhou}}, \bibinfo {author} {\bibfnamefont {Z.}~\bibnamefont {Zhang}}, \emph {et~al.},\ }in\ \href@noop {} {\emph {\bibinfo {booktitle} {Forty-second International Conference on Machine Learning}}}\ (\bibinfo {year} {2025})\BibitemShut {NoStop}%
\bibitem [{\citenamefont {Liu}\ \emph {et~al.}(2026)\citenamefont {Liu}, \citenamefont {Wang}, \citenamefont {Zhang}, \citenamefont {Kariyappa}, \citenamefont {Xiang}, \citenamefont {Chen}, \citenamefont {Suh},\ and\ \citenamefont {Xiao}}]{liu2026reasoningbomb}%
  \BibitemOpen
  \bibfield  {author} {\bibinfo {author} {\bibfnamefont {X.}~\bibnamefont {Liu}}, \bibinfo {author} {\bibfnamefont {X.}~\bibnamefont {Wang}}, \bibinfo {author} {\bibfnamefont {Y.}~\bibnamefont {Zhang}}, \bibinfo {author} {\bibfnamefont {S.}~\bibnamefont {Kariyappa}}, \bibinfo {author} {\bibfnamefont {C.}~\bibnamefont {Xiang}}, \bibinfo {author} {\bibfnamefont {M.}~\bibnamefont {Chen}}, \bibinfo {author} {\bibfnamefont {G.~E.}\ \bibnamefont {Suh}},\ and\ \bibinfo {author} {\bibfnamefont {C.}~\bibnamefont {Xiao}},\ }\href@noop {} {\bibfield  {journal} {\bibinfo  {journal} {arXiv preprint arXiv:2602.00154}\ } (\bibinfo {year} {2026})}\BibitemShut {NoStop}%
\bibitem [{\citenamefont {Schwarzschild}\ \emph {et~al.}(2021)\citenamefont {Schwarzschild}, \citenamefont {Borgnia}, \citenamefont {Gupta}, \citenamefont {Huang}, \citenamefont {Vishkin}, \citenamefont {Goldblum},\ and\ \citenamefont {Goldstein}}]{schwarzschild2021can}%
  \BibitemOpen
  \bibfield  {author} {\bibinfo {author} {\bibfnamefont {A.}~\bibnamefont {Schwarzschild}}, \bibinfo {author} {\bibfnamefont {E.}~\bibnamefont {Borgnia}}, \bibinfo {author} {\bibfnamefont {A.}~\bibnamefont {Gupta}}, \bibinfo {author} {\bibfnamefont {F.}~\bibnamefont {Huang}}, \bibinfo {author} {\bibfnamefont {U.}~\bibnamefont {Vishkin}}, \bibinfo {author} {\bibfnamefont {M.}~\bibnamefont {Goldblum}},\ and\ \bibinfo {author} {\bibfnamefont {T.}~\bibnamefont {Goldstein}},\ }\href@noop {} {\bibfield  {journal} {\bibinfo  {journal} {Advances in Neural Information Processing Systems}\ }\textbf {\bibinfo {volume} {34}},\ \bibinfo {pages} {6695} (\bibinfo {year} {2021})}\BibitemShut {NoStop}%
\bibitem [{\citenamefont {Geiping}\ \emph {et~al.}(2025)\citenamefont {Geiping}, \citenamefont {McLeish}, \citenamefont {Jain}, \citenamefont {Kirchenbauer}, \citenamefont {Singh}, \citenamefont {Bartoldson}, \citenamefont {Kailkhura}, \citenamefont {Bhatele},\ and\ \citenamefont {Goldstein}}]{geiping2025scaling}%
  \BibitemOpen
  \bibfield  {author} {\bibinfo {author} {\bibfnamefont {J.}~\bibnamefont {Geiping}}, \bibinfo {author} {\bibfnamefont {S.~M.}\ \bibnamefont {McLeish}}, \bibinfo {author} {\bibfnamefont {N.}~\bibnamefont {Jain}}, \bibinfo {author} {\bibfnamefont {J.}~\bibnamefont {Kirchenbauer}}, \bibinfo {author} {\bibfnamefont {S.}~\bibnamefont {Singh}}, \bibinfo {author} {\bibfnamefont {B.~R.}\ \bibnamefont {Bartoldson}}, \bibinfo {author} {\bibfnamefont {B.}~\bibnamefont {Kailkhura}}, \bibinfo {author} {\bibfnamefont {A.}~\bibnamefont {Bhatele}},\ and\ \bibinfo {author} {\bibfnamefont {T.}~\bibnamefont {Goldstein}},\ }in\ \href {https://openreview.net/forum?id=S3GhJooWIC} {\emph {\bibinfo {booktitle} {The Thirty-ninth Annual Conference on Neural Information Processing Systems}}}\ (\bibinfo {year} {2025})\BibitemShut {NoStop}%
\bibitem [{\citenamefont {Wang}\ \emph {et~al.}(2023)\citenamefont {Wang}, \citenamefont {Wei}, \citenamefont {Schuurmans}, \citenamefont {Le}, \citenamefont {Chi}, \citenamefont {Narang}, \citenamefont {Chowdhery},\ and\ \citenamefont {Zhou}}]{wang2023selfconsistency}%
  \BibitemOpen
  \bibfield  {author} {\bibinfo {author} {\bibfnamefont {X.}~\bibnamefont {Wang}}, \bibinfo {author} {\bibfnamefont {J.}~\bibnamefont {Wei}}, \bibinfo {author} {\bibfnamefont {D.}~\bibnamefont {Schuurmans}}, \bibinfo {author} {\bibfnamefont {Q.~V.}\ \bibnamefont {Le}}, \bibinfo {author} {\bibfnamefont {E.~H.}\ \bibnamefont {Chi}}, \bibinfo {author} {\bibfnamefont {S.}~\bibnamefont {Narang}}, \bibinfo {author} {\bibfnamefont {A.}~\bibnamefont {Chowdhery}},\ and\ \bibinfo {author} {\bibfnamefont {D.}~\bibnamefont {Zhou}},\ }in\ \href {https://openreview.net/forum?id=1PL1NIMMrw} {\emph {\bibinfo {booktitle} {The Eleventh International Conference on Learning Representations}}}\ (\bibinfo {year} {2023})\BibitemShut {NoStop}%
\bibitem [{\citenamefont {Shojaee}\ \emph {et~al.}(2026)\citenamefont {Shojaee}, \citenamefont {Mirzadeh}, \citenamefont {Horton}, \citenamefont {Bengio}, \citenamefont {Farajtabar} \emph {et~al.}}]{shojaee2026illusion}%
  \BibitemOpen
  \bibfield  {author} {\bibinfo {author} {\bibfnamefont {P.}~\bibnamefont {Shojaee}}, \bibinfo {author} {\bibfnamefont {I.}~\bibnamefont {Mirzadeh}}, \bibinfo {author} {\bibfnamefont {M.}~\bibnamefont {Horton}}, \bibinfo {author} {\bibfnamefont {S.}~\bibnamefont {Bengio}}, \bibinfo {author} {\bibfnamefont {M.}~\bibnamefont {Farajtabar}}, \emph {et~al.},\ }\href@noop {} {\bibfield  {journal} {\bibinfo  {journal} {Advances in Neural Information Processing Systems}\ }\textbf {\bibinfo {volume} {38}},\ \bibinfo {pages} {108018} (\bibinfo {year} {2026})}\BibitemShut {NoStop}%
\bibitem [{\citenamefont {{Center for AI Safety, Scale AI, and HLE Contributors Consortium}}(2026)}]{center2026benchmark}%
  \BibitemOpen
  \bibfield  {author} {\bibinfo {author} {\bibnamefont {{Center for AI Safety, Scale AI, and HLE Contributors Consortium}}},\ }\href@noop {} {\bibfield  {journal} {\bibinfo  {journal} {Nature}\ }\textbf {\bibinfo {volume} {649}},\ \bibinfo {pages} {1139} (\bibinfo {year} {2026})}\BibitemShut {NoStop}%
\bibitem [{\citenamefont {Yang}\ \emph {et~al.}(2026)\citenamefont {Yang}, \citenamefont {Ma}, \citenamefont {Lin},\ and\ \citenamefont {Wei}}]{yang2026towards}%
  \BibitemOpen
  \bibfield  {author} {\bibinfo {author} {\bibfnamefont {W.}~\bibnamefont {Yang}}, \bibinfo {author} {\bibfnamefont {S.}~\bibnamefont {Ma}}, \bibinfo {author} {\bibfnamefont {Y.}~\bibnamefont {Lin}},\ and\ \bibinfo {author} {\bibfnamefont {F.}~\bibnamefont {Wei}},\ }\href@noop {} {\bibfield  {journal} {\bibinfo  {journal} {Advances in Neural Information Processing Systems}\ }\textbf {\bibinfo {volume} {38}},\ \bibinfo {pages} {43605} (\bibinfo {year} {2026})}\BibitemShut {NoStop}%
\bibitem [{\citenamefont {Movahedi}\ \emph {et~al.}(2026)\citenamefont {Movahedi}, \citenamefont {Milovanović}, \citenamefont {Feigin}, \citenamefont {Theus}, \citenamefont {Hofmann}, \citenamefont {Boeva}, \citenamefont {Rusch},\ and\ \citenamefont {Orvieto}}]{movahedi2026fixedpointreasonersstableadaptive}%
  \BibitemOpen
  \bibfield  {author} {\bibinfo {author} {\bibfnamefont {S.}~\bibnamefont {Movahedi}}, \bibinfo {author} {\bibfnamefont {V.}~\bibnamefont {Milovanović}}, \bibinfo {author} {\bibfnamefont {S.~L.}\ \bibnamefont {Feigin}}, \bibinfo {author} {\bibfnamefont {A.}~\bibnamefont {Theus}}, \bibinfo {author} {\bibfnamefont {T.}~\bibnamefont {Hofmann}}, \bibinfo {author} {\bibfnamefont {V.}~\bibnamefont {Boeva}}, \bibinfo {author} {\bibfnamefont {T.~K.}\ \bibnamefont {Rusch}},\ and\ \bibinfo {author} {\bibfnamefont {A.}~\bibnamefont {Orvieto}},\ }\href {https://arxiv.org/abs/2606.18206} {\bibinfo {title} {Fixed-point reasoners: Stable and adaptive deep looped transformers}} (\bibinfo {year} {2026}),\ \Eprint {https://arxiv.org/abs/2606.18206} {arXiv:2606.18206 [cs.AI]} \BibitemShut {NoStop}%
\bibitem [{\citenamefont {Huang}\ \emph {et~al.}(2026)\citenamefont {Huang}, \citenamefont {Geng},\ and\ \citenamefont {Kolter}}]{huang2026equilibrium}%
  \BibitemOpen
  \bibfield  {author} {\bibinfo {author} {\bibfnamefont {B.}~\bibnamefont {Huang}}, \bibinfo {author} {\bibfnamefont {Z.}~\bibnamefont {Geng}},\ and\ \bibinfo {author} {\bibfnamefont {J.~Z.}\ \bibnamefont {Kolter}},\ }in\ \href {https://openreview.net/forum?id=lh95PnOlpM} {\emph {\bibinfo {booktitle} {Forty-third International Conference on Machine Learning}}}\ (\bibinfo {year} {2026})\BibitemShut {NoStop}%
\bibitem [{\citenamefont {Sussillo}\ and\ \citenamefont {Barak}(2013)}]{sussillo2013opening}%
  \BibitemOpen
  \bibfield  {author} {\bibinfo {author} {\bibfnamefont {D.}~\bibnamefont {Sussillo}}\ and\ \bibinfo {author} {\bibfnamefont {O.}~\bibnamefont {Barak}},\ }\href@noop {} {\bibfield  {journal} {\bibinfo  {journal} {Neural computation}\ }\textbf {\bibinfo {volume} {25}},\ \bibinfo {pages} {626} (\bibinfo {year} {2013})}\BibitemShut {NoStop}%
\bibitem [{\citenamefont {Khona}\ and\ \citenamefont {Fiete}(2022)}]{khona2022attractor}%
  \BibitemOpen
  \bibfield  {author} {\bibinfo {author} {\bibfnamefont {M.}~\bibnamefont {Khona}}\ and\ \bibinfo {author} {\bibfnamefont {I.~R.}\ \bibnamefont {Fiete}},\ }\href@noop {} {\bibfield  {journal} {\bibinfo  {journal} {Nature Reviews Neuroscience}\ }\textbf {\bibinfo {volume} {23}},\ \bibinfo {pages} {744} (\bibinfo {year} {2022})}\BibitemShut {NoStop}%
\bibitem [{\citenamefont {Durstewitz}\ \emph {et~al.}(2023)\citenamefont {Durstewitz}, \citenamefont {Koppe},\ and\ \citenamefont {Thurm}}]{durstewitz2023reconstructing}%
  \BibitemOpen
  \bibfield  {author} {\bibinfo {author} {\bibfnamefont {D.}~\bibnamefont {Durstewitz}}, \bibinfo {author} {\bibfnamefont {G.}~\bibnamefont {Koppe}},\ and\ \bibinfo {author} {\bibfnamefont {M.~I.}\ \bibnamefont {Thurm}},\ }\href@noop {} {\bibfield  {journal} {\bibinfo  {journal} {Nature Reviews Neuroscience}\ }\textbf {\bibinfo {volume} {24}},\ \bibinfo {pages} {693} (\bibinfo {year} {2023})}\BibitemShut {NoStop}%
\bibitem [{\citenamefont {Prairie}\ \emph {et~al.}(2026)\citenamefont {Prairie}, \citenamefont {Novack}, \citenamefont {Berg-Kirkpatrick},\ and\ \citenamefont {Fu}}]{prairie2026parcaescalinglawsstable}%
  \BibitemOpen
  \bibfield  {author} {\bibinfo {author} {\bibfnamefont {H.}~\bibnamefont {Prairie}}, \bibinfo {author} {\bibfnamefont {Z.}~\bibnamefont {Novack}}, \bibinfo {author} {\bibfnamefont {T.}~\bibnamefont {Berg-Kirkpatrick}},\ and\ \bibinfo {author} {\bibfnamefont {D.~Y.}\ \bibnamefont {Fu}},\ }\href {https://arxiv.org/abs/2604.12946} {\bibinfo {title} {Parcae: Scaling laws for stable looped language models}} (\bibinfo {year} {2026}),\ \Eprint {https://arxiv.org/abs/2604.12946} {arXiv:2604.12946 [cs.LG]} \BibitemShut {NoStop}%
\bibitem [{\citenamefont {Moehlis}\ \emph {et~al.}(2004)\citenamefont {Moehlis}, \citenamefont {Faisst},\ and\ \citenamefont {Eckhardt}}]{moehlis2004low}%
  \BibitemOpen
  \bibfield  {author} {\bibinfo {author} {\bibfnamefont {J.}~\bibnamefont {Moehlis}}, \bibinfo {author} {\bibfnamefont {H.}~\bibnamefont {Faisst}},\ and\ \bibinfo {author} {\bibfnamefont {B.}~\bibnamefont {Eckhardt}},\ }\href@noop {} {\bibfield  {journal} {\bibinfo  {journal} {New Journal of Physics}\ }\textbf {\bibinfo {volume} {6}},\ \bibinfo {pages} {56} (\bibinfo {year} {2004})}\BibitemShut {NoStop}%
\bibitem [{\citenamefont {Altmann}\ \emph {et~al.}(2013)\citenamefont {Altmann}, \citenamefont {Portela},\ and\ \citenamefont {T\'el}}]{altmann2013leaking}%
  \BibitemOpen
  \bibfield  {author} {\bibinfo {author} {\bibfnamefont {E.~G.}\ \bibnamefont {Altmann}}, \bibinfo {author} {\bibfnamefont {J.~S.~E.}\ \bibnamefont {Portela}},\ and\ \bibinfo {author} {\bibfnamefont {T.}~\bibnamefont {T\'el}},\ }\href {https://doi.org/10.1103/RevModPhys.85.869} {\bibfield  {journal} {\bibinfo  {journal} {Rev. Mod. Phys.}\ }\textbf {\bibinfo {volume} {85}},\ \bibinfo {pages} {869} (\bibinfo {year} {2013})}\BibitemShut {NoStop}%
\bibitem [{\citenamefont {Hopfield}\ and\ \citenamefont {Tank}(1985)}]{hopfield1985neural}%
  \BibitemOpen
  \bibfield  {author} {\bibinfo {author} {\bibfnamefont {J.~J.}\ \bibnamefont {Hopfield}}\ and\ \bibinfo {author} {\bibfnamefont {D.~W.}\ \bibnamefont {Tank}},\ }\href@noop {} {\bibfield  {journal} {\bibinfo  {journal} {Biological cybernetics}\ }\textbf {\bibinfo {volume} {52}},\ \bibinfo {pages} {141} (\bibinfo {year} {1985})}\BibitemShut {NoStop}%
\bibitem [{\citenamefont {Chen}\ and\ \citenamefont {Aihara}(1995)}]{chen1995chaotic}%
  \BibitemOpen
  \bibfield  {author} {\bibinfo {author} {\bibfnamefont {L.}~\bibnamefont {Chen}}\ and\ \bibinfo {author} {\bibfnamefont {K.}~\bibnamefont {Aihara}},\ }\href@noop {} {\bibfield  {journal} {\bibinfo  {journal} {Neural networks}\ }\textbf {\bibinfo {volume} {8}},\ \bibinfo {pages} {915} (\bibinfo {year} {1995})}\BibitemShut {NoStop}%
\bibitem [{\citenamefont {Elser}\ \emph {et~al.}(2007)\citenamefont {Elser}, \citenamefont {Rankenburg},\ and\ \citenamefont {Thibault}}]{elser2007searching}%
  \BibitemOpen
  \bibfield  {author} {\bibinfo {author} {\bibfnamefont {V.}~\bibnamefont {Elser}}, \bibinfo {author} {\bibfnamefont {I.}~\bibnamefont {Rankenburg}},\ and\ \bibinfo {author} {\bibfnamefont {P.}~\bibnamefont {Thibault}},\ }\href@noop {} {\bibfield  {journal} {\bibinfo  {journal} {Proceedings of the National Academy of Sciences}\ }\textbf {\bibinfo {volume} {104}},\ \bibinfo {pages} {418} (\bibinfo {year} {2007})}\BibitemShut {NoStop}%
\bibitem [{\citenamefont {Ercsey-Ravasz}\ and\ \citenamefont {Toroczkai}(2011)}]{ercsey2011optimization}%
  \BibitemOpen
  \bibfield  {author} {\bibinfo {author} {\bibfnamefont {M.}~\bibnamefont {Ercsey-Ravasz}}\ and\ \bibinfo {author} {\bibfnamefont {Z.}~\bibnamefont {Toroczkai}},\ }\href@noop {} {\bibfield  {journal} {\bibinfo  {journal} {Nature Physics}\ }\textbf {\bibinfo {volume} {7}},\ \bibinfo {pages} {966} (\bibinfo {year} {2011})}\BibitemShut {NoStop}%
\bibitem [{\citenamefont {Ercsey-Ravasz}\ and\ \citenamefont {Toroczkai}(2012)}]{ercsey2012chaos}%
  \BibitemOpen
  \bibfield  {author} {\bibinfo {author} {\bibfnamefont {M.}~\bibnamefont {Ercsey-Ravasz}}\ and\ \bibinfo {author} {\bibfnamefont {Z.}~\bibnamefont {Toroczkai}},\ }\href@noop {} {\bibfield  {journal} {\bibinfo  {journal} {Scientific reports}\ }\textbf {\bibinfo {volume} {2}},\ \bibinfo {pages} {725} (\bibinfo {year} {2012})}\BibitemShut {NoStop}%
\bibitem [{\citenamefont {Daza}\ \emph {et~al.}(2016)\citenamefont {Daza}, \citenamefont {Wagemakers}, \citenamefont {Georgeot}, \citenamefont {Gu{\'e}ry-Odelin},\ and\ \citenamefont {Sanju{\'a}n}}]{daza2016basin}%
  \BibitemOpen
  \bibfield  {author} {\bibinfo {author} {\bibfnamefont {A.}~\bibnamefont {Daza}}, \bibinfo {author} {\bibfnamefont {A.}~\bibnamefont {Wagemakers}}, \bibinfo {author} {\bibfnamefont {B.}~\bibnamefont {Georgeot}}, \bibinfo {author} {\bibfnamefont {D.}~\bibnamefont {Gu{\'e}ry-Odelin}},\ and\ \bibinfo {author} {\bibfnamefont {M.~A.}\ \bibnamefont {Sanju{\'a}n}},\ }\href@noop {} {\bibfield  {journal} {\bibinfo  {journal} {Scientific reports}\ }\textbf {\bibinfo {volume} {6}},\ \bibinfo {pages} {31416} (\bibinfo {year} {2016})}\BibitemShut {NoStop}%
\bibitem [{\citenamefont {Grebogi}\ \emph {et~al.}(1983)\citenamefont {Grebogi}, \citenamefont {Ott},\ and\ \citenamefont {Yorke}}]{grebogi1983fractal}%
  \BibitemOpen
  \bibfield  {author} {\bibinfo {author} {\bibfnamefont {C.}~\bibnamefont {Grebogi}}, \bibinfo {author} {\bibfnamefont {E.}~\bibnamefont {Ott}},\ and\ \bibinfo {author} {\bibfnamefont {J.~A.}\ \bibnamefont {Yorke}},\ }\href@noop {} {\bibfield  {journal} {\bibinfo  {journal} {Physical Review Letters}\ }\textbf {\bibinfo {volume} {50}},\ \bibinfo {pages} {935} (\bibinfo {year} {1983})}\BibitemShut {NoStop}%
\bibitem [{\citenamefont {T{\'e}l}\ and\ \citenamefont {Lai}(2008)}]{tel2008chaotic}%
  \BibitemOpen
  \bibfield  {author} {\bibinfo {author} {\bibfnamefont {T.}~\bibnamefont {T{\'e}l}}\ and\ \bibinfo {author} {\bibfnamefont {Y.-C.}\ \bibnamefont {Lai}},\ }\href@noop {} {\bibfield  {journal} {\bibinfo  {journal} {Physics Reports}\ }\textbf {\bibinfo {volume} {460}},\ \bibinfo {pages} {245} (\bibinfo {year} {2008})}\BibitemShut {NoStop}%
\bibitem [{\citenamefont {Motter}\ \emph {et~al.}(2013)\citenamefont {Motter}, \citenamefont {Gruiz}, \citenamefont {K{\'a}rolyi},\ and\ \citenamefont {T{\'e}l}}]{motter2013doubly}%
  \BibitemOpen
  \bibfield  {author} {\bibinfo {author} {\bibfnamefont {A.~E.}\ \bibnamefont {Motter}}, \bibinfo {author} {\bibfnamefont {M.}~\bibnamefont {Gruiz}}, \bibinfo {author} {\bibfnamefont {G.}~\bibnamefont {K{\'a}rolyi}},\ and\ \bibinfo {author} {\bibfnamefont {T.}~\bibnamefont {T{\'e}l}},\ }\href@noop {} {\bibfield  {journal} {\bibinfo  {journal} {Physical review letters}\ }\textbf {\bibinfo {volume} {111}},\ \bibinfo {pages} {194101} (\bibinfo {year} {2013})}\BibitemShut {NoStop}%
\bibitem [{\citenamefont {Dauphin}\ \emph {et~al.}(2014)\citenamefont {Dauphin}, \citenamefont {Pascanu}, \citenamefont {Gulcehre}, \citenamefont {Cho}, \citenamefont {Ganguli},\ and\ \citenamefont {Bengio}}]{dauphin2014identifying}%
  \BibitemOpen
  \bibfield  {author} {\bibinfo {author} {\bibfnamefont {Y.~N.}\ \bibnamefont {Dauphin}}, \bibinfo {author} {\bibfnamefont {R.}~\bibnamefont {Pascanu}}, \bibinfo {author} {\bibfnamefont {C.}~\bibnamefont {Gulcehre}}, \bibinfo {author} {\bibfnamefont {K.}~\bibnamefont {Cho}}, \bibinfo {author} {\bibfnamefont {S.}~\bibnamefont {Ganguli}},\ and\ \bibinfo {author} {\bibfnamefont {Y.}~\bibnamefont {Bengio}},\ }\href@noop {} {\bibfield  {journal} {\bibinfo  {journal} {Advances in neural information processing systems}\ }\textbf {\bibinfo {volume} {27}} (\bibinfo {year} {2014})}\BibitemShut {NoStop}%
\bibitem [{\citenamefont {Kantz}\ and\ \citenamefont {Grassberger}(1985)}]{kantz1985repellers}%
  \BibitemOpen
  \bibfield  {author} {\bibinfo {author} {\bibfnamefont {H.}~\bibnamefont {Kantz}}\ and\ \bibinfo {author} {\bibfnamefont {P.}~\bibnamefont {Grassberger}},\ }\href@noop {} {\bibfield  {journal} {\bibinfo  {journal} {Physica D: Nonlinear Phenomena}\ }\textbf {\bibinfo {volume} {17}},\ \bibinfo {pages} {75} (\bibinfo {year} {1985})}\BibitemShut {NoStop}%
\bibitem [{\citenamefont {Froeschl{\'e}}\ \emph {et~al.}(1997)\citenamefont {Froeschl{\'e}}, \citenamefont {Gonczi},\ and\ \citenamefont {Lega}}]{froeschle1997fast}%
  \BibitemOpen
  \bibfield  {author} {\bibinfo {author} {\bibfnamefont {C.}~\bibnamefont {Froeschl{\'e}}}, \bibinfo {author} {\bibfnamefont {R.}~\bibnamefont {Gonczi}},\ and\ \bibinfo {author} {\bibfnamefont {E.}~\bibnamefont {Lega}},\ }\href@noop {} {\bibfield  {journal} {\bibinfo  {journal} {Planetary and space science}\ }\textbf {\bibinfo {volume} {45}},\ \bibinfo {pages} {881} (\bibinfo {year} {1997})}\BibitemShut {NoStop}%
\bibitem [{\citenamefont {Ben~Arous}\ \emph {et~al.}(2021)\citenamefont {Ben~Arous}, \citenamefont {Fyodorov},\ and\ \citenamefont {Khoruzhenko}}]{ben2021counting}%
  \BibitemOpen
  \bibfield  {author} {\bibinfo {author} {\bibfnamefont {G.}~\bibnamefont {Ben~Arous}}, \bibinfo {author} {\bibfnamefont {Y.~V.}\ \bibnamefont {Fyodorov}},\ and\ \bibinfo {author} {\bibfnamefont {B.~A.}\ \bibnamefont {Khoruzhenko}},\ }\href@noop {} {\bibfield  {journal} {\bibinfo  {journal} {Proceedings of the National Academy of Sciences}\ }\textbf {\bibinfo {volume} {118}},\ \bibinfo {pages} {e2023719118} (\bibinfo {year} {2021})}\BibitemShut {NoStop}%
\bibitem [{\citenamefont {Wainrib}\ and\ \citenamefont {Touboul}(2013)}]{wainrib2013topological}%
  \BibitemOpen
  \bibfield  {author} {\bibinfo {author} {\bibfnamefont {G.}~\bibnamefont {Wainrib}}\ and\ \bibinfo {author} {\bibfnamefont {J.}~\bibnamefont {Touboul}},\ }\href@noop {} {\bibfield  {journal} {\bibinfo  {journal} {Physical review letters}\ }\textbf {\bibinfo {volume} {110}},\ \bibinfo {pages} {118101} (\bibinfo {year} {2013})}\BibitemShut {NoStop}%
\bibitem [{\citenamefont {Dechter}\ and\ \citenamefont {Meiri}(1994)}]{dechter1994experimental}%
  \BibitemOpen
  \bibfield  {author} {\bibinfo {author} {\bibfnamefont {R.}~\bibnamefont {Dechter}}\ and\ \bibinfo {author} {\bibfnamefont {I.}~\bibnamefont {Meiri}},\ }\href@noop {} {\bibfield  {journal} {\bibinfo  {journal} {Artificial Intelligence}\ }\textbf {\bibinfo {volume} {68}},\ \bibinfo {pages} {211} (\bibinfo {year} {1994})}\BibitemShut {NoStop}%
\bibitem [{\citenamefont {Cheeseman}\ \emph {et~al.}(1991)\citenamefont {Cheeseman}, \citenamefont {Kanefsky}, \citenamefont {Taylor} \emph {et~al.}}]{cheeseman1991really}%
  \BibitemOpen
  \bibfield  {author} {\bibinfo {author} {\bibfnamefont {P.~C.}\ \bibnamefont {Cheeseman}}, \bibinfo {author} {\bibfnamefont {B.}~\bibnamefont {Kanefsky}}, \bibinfo {author} {\bibfnamefont {W.~M.}\ \bibnamefont {Taylor}}, \emph {et~al.},\ }in\ \href@noop {} {\emph {\bibinfo {booktitle} {Ijcai}}},\ Vol.~\bibinfo {volume} {91}\ (\bibinfo {year} {1991})\ pp.\ \bibinfo {pages} {331--337}\BibitemShut {NoStop}%
\bibitem [{\citenamefont {Monasson}\ \emph {et~al.}(1999)\citenamefont {Monasson}, \citenamefont {Zecchina}, \citenamefont {Kirkpatrick}, \citenamefont {Selman},\ and\ \citenamefont {Troyansky}}]{monasson1999determining}%
  \BibitemOpen
  \bibfield  {author} {\bibinfo {author} {\bibfnamefont {R.}~\bibnamefont {Monasson}}, \bibinfo {author} {\bibfnamefont {R.}~\bibnamefont {Zecchina}}, \bibinfo {author} {\bibfnamefont {S.}~\bibnamefont {Kirkpatrick}}, \bibinfo {author} {\bibfnamefont {B.}~\bibnamefont {Selman}},\ and\ \bibinfo {author} {\bibfnamefont {L.}~\bibnamefont {Troyansky}},\ }\href@noop {} {\bibfield  {journal} {\bibinfo  {journal} {Nature}\ }\textbf {\bibinfo {volume} {400}},\ \bibinfo {pages} {133} (\bibinfo {year} {1999})}\BibitemShut {NoStop}%
\bibitem [{\citenamefont {Lai}\ and\ \citenamefont {T{\'e}l}(2011)}]{lai2011transient}%
  \BibitemOpen
  \bibfield  {author} {\bibinfo {author} {\bibfnamefont {Y.-C.}\ \bibnamefont {Lai}}\ and\ \bibinfo {author} {\bibfnamefont {T.}~\bibnamefont {T{\'e}l}},\ }\href@noop {} {\emph {\bibinfo {title} {Transient chaos: complex dynamics on finite time scales}}}\ (\bibinfo  {publisher} {Springer Science \& Business Media},\ \bibinfo {year} {2011})\BibitemShut {NoStop}%
\bibitem [{\citenamefont {Braunstein}\ \emph {et~al.}(2002)\citenamefont {Braunstein}, \citenamefont {Leone}, \citenamefont {Ricci-Tersenghi},\ and\ \citenamefont {Zecchina}}]{braunstein2002complexity}%
  \BibitemOpen
  \bibfield  {author} {\bibinfo {author} {\bibfnamefont {A.}~\bibnamefont {Braunstein}}, \bibinfo {author} {\bibfnamefont {M.}~\bibnamefont {Leone}}, \bibinfo {author} {\bibfnamefont {F.}~\bibnamefont {Ricci-Tersenghi}},\ and\ \bibinfo {author} {\bibfnamefont {R.}~\bibnamefont {Zecchina}},\ }\href@noop {} {\bibfield  {journal} {\bibinfo  {journal} {Journal of Physics A: Mathematical and General}\ }\textbf {\bibinfo {volume} {35}},\ \bibinfo {pages} {7559} (\bibinfo {year} {2002})}\BibitemShut {NoStop}%
\bibitem [{\citenamefont {Moore}(1990)}]{moore1990unpredictability}%
  \BibitemOpen
  \bibfield  {author} {\bibinfo {author} {\bibfnamefont {C.}~\bibnamefont {Moore}},\ }\href@noop {} {\bibfield  {journal} {\bibinfo  {journal} {Physical Review Letters}\ }\textbf {\bibinfo {volume} {64}},\ \bibinfo {pages} {2354} (\bibinfo {year} {1990})}\BibitemShut {NoStop}%
\bibitem [{\citenamefont {Shaw}(1981)}]{shaw1981strange}%
  \BibitemOpen
  \bibfield  {author} {\bibinfo {author} {\bibfnamefont {R.}~\bibnamefont {Shaw}},\ }\href@noop {} {\bibfield  {journal} {\bibinfo  {journal} {Zeitschrift f{\"u}r Naturforschung A}\ }\textbf {\bibinfo {volume} {36}},\ \bibinfo {pages} {80} (\bibinfo {year} {1981})}\BibitemShut {NoStop}%
\bibitem [{\citenamefont {Crutchfield}\ and\ \citenamefont {Young}(1989)}]{crutchfield1989inferring}%
  \BibitemOpen
  \bibfield  {author} {\bibinfo {author} {\bibfnamefont {J.~P.}\ \bibnamefont {Crutchfield}}\ and\ \bibinfo {author} {\bibfnamefont {K.}~\bibnamefont {Young}},\ }\href@noop {} {\bibfield  {journal} {\bibinfo  {journal} {Physical Review Letters}\ }\textbf {\bibinfo {volume} {63}},\ \bibinfo {pages} {105} (\bibinfo {year} {1989})}\BibitemShut {NoStop}%
\end{thebibliography}%


\begin{thebibliography}{15}%
\makeatletter
\providecommand \@ifxundefined [1]{%
 \@ifx{#1\undefined}
}%
\providecommand \@ifnum [1]{%
 \ifnum #1\expandafter \@firstoftwo
 \else \expandafter \@secondoftwo
 \fi
}%
\providecommand \@ifx [1]{%
 \ifx #1\expandafter \@firstoftwo
 \else \expandafter \@secondoftwo
 \fi
}%
\providecommand \natexlab [1]{#1}%
\providecommand \enquote  [1]{``#1''}%
\providecommand \bibnamefont  [1]{#1}%
\providecommand \bibfnamefont [1]{#1}%
\providecommand \citenamefont [1]{#1}%
\providecommand \href@noop [0]{\@secondoftwo}%
\providecommand \href [0]{\begingroup \@sanitize@url \@href}%
\providecommand \@href[1]{\@@startlink{#1}\@@href}%
\providecommand \@@href[1]{\endgroup#1\@@endlink}%
\providecommand \@sanitize@url [0]{\catcode `\\12\catcode `\$12\catcode `\&12\catcode `\#12\catcode `\^12\catcode `\_12\catcode `\%12\relax}%
\providecommand \@@startlink[1]{}%
\providecommand \@@endlink[0]{}%
\providecommand \url  [0]{\begingroup\@sanitize@url \@url }%
\providecommand \@url [1]{\endgroup\@href {#1}{\urlprefix }}%
\providecommand \urlprefix  [0]{URL }%
\providecommand \Eprint [0]{\href }%
\providecommand \doibase [0]{https://doi.org/}%
\providecommand \selectlanguage [0]{\@gobble}%
\providecommand \bibinfo  [0]{\@secondoftwo}%
\providecommand \bibfield  [0]{\@secondoftwo}%
\providecommand \translation [1]{[#1]}%
\providecommand \BibitemOpen [0]{}%
\providecommand \bibitemStop [0]{}%
\providecommand \bibitemNoStop [0]{.\EOS\space}%
\providecommand \EOS [0]{\spacefactor3000\relax}%
\providecommand \BibitemShut  [1]{\csname bibitem#1\endcsname}%
\let\auto@bib@innerbib\@empty
\bibitem [{\citenamefont {Huang}\ \emph {et~al.}(2026)\citenamefont {Huang}, \citenamefont {Geng},\ and\ \citenamefont {Kolter}}]{huang2026equilibrium}%
  \BibitemOpen
  \bibfield  {author} {\bibinfo {author} {\bibfnamefont {B.}~\bibnamefont {Huang}}, \bibinfo {author} {\bibfnamefont {Z.}~\bibnamefont {Geng}},\ and\ \bibinfo {author} {\bibfnamefont {J.~Z.}\ \bibnamefont {Kolter}},\ }in\ \href {https://openreview.net/forum?id=lh95PnOlpM} {\emph {\bibinfo {booktitle} {Forty-third International Conference on Machine Learning}}}\ (\bibinfo {year} {2026})\BibitemShut {NoStop}%
\bibitem [{\citenamefont {Movahedi}\ \emph {et~al.}(2026)\citenamefont {Movahedi}, \citenamefont {Milovanović}, \citenamefont {Feigin}, \citenamefont {Theus}, \citenamefont {Hofmann}, \citenamefont {Boeva}, \citenamefont {Rusch},\ and\ \citenamefont {Orvieto}}]{movahedi2026fixedpointreasonersstableadaptive}%
  \BibitemOpen
  \bibfield  {author} {\bibinfo {author} {\bibfnamefont {S.}~\bibnamefont {Movahedi}}, \bibinfo {author} {\bibfnamefont {V.}~\bibnamefont {Milovanović}}, \bibinfo {author} {\bibfnamefont {S.~L.}\ \bibnamefont {Feigin}}, \bibinfo {author} {\bibfnamefont {A.}~\bibnamefont {Theus}}, \bibinfo {author} {\bibfnamefont {T.}~\bibnamefont {Hofmann}}, \bibinfo {author} {\bibfnamefont {V.}~\bibnamefont {Boeva}}, \bibinfo {author} {\bibfnamefont {T.~K.}\ \bibnamefont {Rusch}},\ and\ \bibinfo {author} {\bibfnamefont {A.}~\bibnamefont {Orvieto}},\ }\href {https://arxiv.org/abs/2606.18206} {\bibinfo {title} {Fixed-point reasoners: Stable and adaptive deep looped transformers}} (\bibinfo {year} {2026}),\ \Eprint {https://arxiv.org/abs/2606.18206} {arXiv:2606.18206 [cs.AI]} \BibitemShut {NoStop}%
\bibitem [{\citenamefont {Prairie}\ \emph {et~al.}(2026)\citenamefont {Prairie}, \citenamefont {Novack}, \citenamefont {Berg-Kirkpatrick},\ and\ \citenamefont {Fu}}]{prairie2026parcaescalinglawsstable}%
  \BibitemOpen
  \bibfield  {author} {\bibinfo {author} {\bibfnamefont {H.}~\bibnamefont {Prairie}}, \bibinfo {author} {\bibfnamefont {Z.}~\bibnamefont {Novack}}, \bibinfo {author} {\bibfnamefont {T.}~\bibnamefont {Berg-Kirkpatrick}},\ and\ \bibinfo {author} {\bibfnamefont {D.~Y.}\ \bibnamefont {Fu}},\ }\href {https://arxiv.org/abs/2604.12946} {\bibinfo {title} {Parcae: Scaling laws for stable looped language models}} (\bibinfo {year} {2026}),\ \Eprint {https://arxiv.org/abs/2604.12946} {arXiv:2604.12946 [cs.LG]} \BibitemShut {NoStop}%
\bibitem [{\citenamefont {Wang}\ \emph {et~al.}(2025)\citenamefont {Wang}, \citenamefont {Li}, \citenamefont {Sun}, \citenamefont {Chen}, \citenamefont {Liu}, \citenamefont {Wu}, \citenamefont {Lu}, \citenamefont {Song},\ and\ \citenamefont {Yadkori}}]{wang2025hierarchicalreasoningmodel}%
  \BibitemOpen
  \bibfield  {author} {\bibinfo {author} {\bibfnamefont {G.}~\bibnamefont {Wang}}, \bibinfo {author} {\bibfnamefont {J.}~\bibnamefont {Li}}, \bibinfo {author} {\bibfnamefont {Y.}~\bibnamefont {Sun}}, \bibinfo {author} {\bibfnamefont {X.}~\bibnamefont {Chen}}, \bibinfo {author} {\bibfnamefont {C.}~\bibnamefont {Liu}}, \bibinfo {author} {\bibfnamefont {Y.}~\bibnamefont {Wu}}, \bibinfo {author} {\bibfnamefont {M.}~\bibnamefont {Lu}}, \bibinfo {author} {\bibfnamefont {S.}~\bibnamefont {Song}},\ and\ \bibinfo {author} {\bibfnamefont {Y.~A.}\ \bibnamefont {Yadkori}},\ }\href {https://arxiv.org/abs/2506.21734} {\bibinfo {title} {Hierarchical reasoning model}} (\bibinfo {year} {2025}),\ \Eprint {https://arxiv.org/abs/2506.21734} {arXiv:2506.21734 [cs.AI]} \BibitemShut {NoStop}%
\bibitem [{\citenamefont {Daza}\ \emph {et~al.}(2016)\citenamefont {Daza}, \citenamefont {Wagemakers}, \citenamefont {Georgeot}, \citenamefont {Gu{\'e}ry-Odelin},\ and\ \citenamefont {Sanju{\'a}n}}]{daza2016basin}%
  \BibitemOpen
  \bibfield  {author} {\bibinfo {author} {\bibfnamefont {A.}~\bibnamefont {Daza}}, \bibinfo {author} {\bibfnamefont {A.}~\bibnamefont {Wagemakers}}, \bibinfo {author} {\bibfnamefont {B.}~\bibnamefont {Georgeot}}, \bibinfo {author} {\bibfnamefont {D.}~\bibnamefont {Gu{\'e}ry-Odelin}},\ and\ \bibinfo {author} {\bibfnamefont {M.~A.}\ \bibnamefont {Sanju{\'a}n}},\ }\href@noop {} {\bibfield  {journal} {\bibinfo  {journal} {Scientific reports}\ }\textbf {\bibinfo {volume} {6}},\ \bibinfo {pages} {31416} (\bibinfo {year} {2016})}\BibitemShut {NoStop}%
\bibitem [{\citenamefont {McDonald}\ \emph {et~al.}(1985)\citenamefont {McDonald}, \citenamefont {Grebogi}, \citenamefont {Ott},\ and\ \citenamefont {Yorke}}]{mcdonald1985fractal}%
  \BibitemOpen
  \bibfield  {author} {\bibinfo {author} {\bibfnamefont {S.~W.}\ \bibnamefont {McDonald}}, \bibinfo {author} {\bibfnamefont {C.}~\bibnamefont {Grebogi}}, \bibinfo {author} {\bibfnamefont {E.}~\bibnamefont {Ott}},\ and\ \bibinfo {author} {\bibfnamefont {J.~A.}\ \bibnamefont {Yorke}},\ }\href@noop {} {\bibfield  {journal} {\bibinfo  {journal} {Physica D: Nonlinear Phenomena}\ }\textbf {\bibinfo {volume} {17}},\ \bibinfo {pages} {125} (\bibinfo {year} {1985})}\BibitemShut {NoStop}%
\bibitem [{\citenamefont {Froeschl{\'e}}\ \emph {et~al.}(1997)\citenamefont {Froeschl{\'e}}, \citenamefont {Gonczi},\ and\ \citenamefont {Lega}}]{froeschle1997fast}%
  \BibitemOpen
  \bibfield  {author} {\bibinfo {author} {\bibfnamefont {C.}~\bibnamefont {Froeschl{\'e}}}, \bibinfo {author} {\bibfnamefont {R.}~\bibnamefont {Gonczi}},\ and\ \bibinfo {author} {\bibfnamefont {E.}~\bibnamefont {Lega}},\ }\href@noop {} {\bibfield  {journal} {\bibinfo  {journal} {Planetary and space science}\ }\textbf {\bibinfo {volume} {45}},\ \bibinfo {pages} {881} (\bibinfo {year} {1997})}\BibitemShut {NoStop}%
\bibitem [{\citenamefont {Grebogi}\ \emph {et~al.}(1983)\citenamefont {Grebogi}, \citenamefont {McDonald}, \citenamefont {Ott},\ and\ \citenamefont {Yorke}}]{grebogi1983final}%
  \BibitemOpen
  \bibfield  {author} {\bibinfo {author} {\bibfnamefont {C.}~\bibnamefont {Grebogi}}, \bibinfo {author} {\bibfnamefont {S.~W.}\ \bibnamefont {McDonald}}, \bibinfo {author} {\bibfnamefont {E.}~\bibnamefont {Ott}},\ and\ \bibinfo {author} {\bibfnamefont {J.~A.}\ \bibnamefont {Yorke}},\ }\href@noop {} {\bibfield  {journal} {\bibinfo  {journal} {Physics Letters A}\ }\textbf {\bibinfo {volume} {99}},\ \bibinfo {pages} {415} (\bibinfo {year} {1983})}\BibitemShut {NoStop}%
\bibitem [{\citenamefont {Mandelbrot}(1967)}]{mandelbrot1967long}%
  \BibitemOpen
  \bibfield  {author} {\bibinfo {author} {\bibfnamefont {B.}~\bibnamefont {Mandelbrot}},\ }\href@noop {} {\bibfield  {journal} {\bibinfo  {journal} {science}\ }\textbf {\bibinfo {volume} {156}},\ \bibinfo {pages} {636} (\bibinfo {year} {1967})}\BibitemShut {NoStop}%
\bibitem [{\citenamefont {Chen}\ \emph {et~al.}(2017)\citenamefont {Chen}, \citenamefont {Nishikawa},\ and\ \citenamefont {Motter}}]{chen2017slim}%
  \BibitemOpen
  \bibfield  {author} {\bibinfo {author} {\bibfnamefont {X.}~\bibnamefont {Chen}}, \bibinfo {author} {\bibfnamefont {T.}~\bibnamefont {Nishikawa}},\ and\ \bibinfo {author} {\bibfnamefont {A.~E.}\ \bibnamefont {Motter}},\ }\href@noop {} {\bibfield  {journal} {\bibinfo  {journal} {Physical Review X}\ }\textbf {\bibinfo {volume} {7}},\ \bibinfo {pages} {021040} (\bibinfo {year} {2017})}\BibitemShut {NoStop}%
\bibitem [{\citenamefont {Motter}\ \emph {et~al.}(2013)\citenamefont {Motter}, \citenamefont {Gruiz}, \citenamefont {K{\'a}rolyi},\ and\ \citenamefont {T{\'e}l}}]{motter2013doubly}%
  \BibitemOpen
  \bibfield  {author} {\bibinfo {author} {\bibfnamefont {A.~E.}\ \bibnamefont {Motter}}, \bibinfo {author} {\bibfnamefont {M.}~\bibnamefont {Gruiz}}, \bibinfo {author} {\bibfnamefont {G.}~\bibnamefont {K{\'a}rolyi}},\ and\ \bibinfo {author} {\bibfnamefont {T.}~\bibnamefont {T{\'e}l}},\ }\href@noop {} {\bibfield  {journal} {\bibinfo  {journal} {Physical review letters}\ }\textbf {\bibinfo {volume} {111}},\ \bibinfo {pages} {194101} (\bibinfo {year} {2013})}\BibitemShut {NoStop}%
\bibitem [{\citenamefont {Omel’chenko}\ and\ \citenamefont {Tel}(2022)}]{omel2022focusing}%
  \BibitemOpen
  \bibfield  {author} {\bibinfo {author} {\bibfnamefont {O.~E.}\ \bibnamefont {Omel’chenko}}\ and\ \bibinfo {author} {\bibfnamefont {T.}~\bibnamefont {Tel}},\ }\href@noop {} {\bibfield  {journal} {\bibinfo  {journal} {Journal of Physics: Complexity}\ }\textbf {\bibinfo {volume} {3}},\ \bibinfo {pages} {010201} (\bibinfo {year} {2022})}\BibitemShut {NoStop}%
\bibitem [{\citenamefont {K{\'a}rolyi}\ and\ \citenamefont {T{\'e}l}(2021)}]{karolyi2021new}%
  \BibitemOpen
  \bibfield  {author} {\bibinfo {author} {\bibfnamefont {G.}~\bibnamefont {K{\'a}rolyi}}\ and\ \bibinfo {author} {\bibfnamefont {T.}~\bibnamefont {T{\'e}l}},\ }\href@noop {} {\bibfield  {journal} {\bibinfo  {journal} {Journal of Physics: Complexity}\ }\textbf {\bibinfo {volume} {2}},\ \bibinfo {pages} {035001} (\bibinfo {year} {2021})}\BibitemShut {NoStop}%
\bibitem [{\citenamefont {Stone}\ and\ \citenamefont {Leigh}(2019)}]{stone2019statistical}%
  \BibitemOpen
  \bibfield  {author} {\bibinfo {author} {\bibfnamefont {N.~C.}\ \bibnamefont {Stone}}\ and\ \bibinfo {author} {\bibfnamefont {N.~W.}\ \bibnamefont {Leigh}},\ }\href@noop {} {\bibfield  {journal} {\bibinfo  {journal} {Nature}\ }\textbf {\bibinfo {volume} {576}},\ \bibinfo {pages} {406} (\bibinfo {year} {2019})}\BibitemShut {NoStop}%
\bibitem [{\citenamefont {Wiggins}(1994)}]{wiggins1994normally}%
  \BibitemOpen
  \bibfield  {author} {\bibinfo {author} {\bibfnamefont {S.}~\bibnamefont {Wiggins}},\ }\href@noop {} {\emph {\bibinfo {title} {Normally hyperbolic invariant manifolds in dynamical systems}}},\ Vol.\ \bibinfo {volume} {105}\ (\bibinfo  {publisher} {Springer Science \& Business Media},\ \bibinfo {year} {1994})\BibitemShut {NoStop}%
\end{thebibliography}%


\begin{thebibliography}{0}%
\makeatletter
\providecommand \@ifxundefined [1]{%
 \@ifx{#1\undefined}
}%
\providecommand \@ifnum [1]{%
 \ifnum #1\expandafter \@firstoftwo
 \else \expandafter \@secondoftwo
 \fi
}%
\providecommand \@ifx [1]{%
 \ifx #1\expandafter \@firstoftwo
 \else \expandafter \@secondoftwo
 \fi
}%
\providecommand \natexlab [1]{#1}%
\providecommand \enquote  [1]{``#1''}%
\providecommand \bibnamefont  [1]{#1}%
\providecommand \bibfnamefont [1]{#1}%
\providecommand \citenamefont [1]{#1}%
\providecommand \href@noop [0]{\@secondoftwo}%
\providecommand \href [0]{\begingroup \@sanitize@url \@href}%
\providecommand \@href[1]{\@@startlink{#1}\@@href}%
\providecommand \@@href[1]{\endgroup#1\@@endlink}%
\providecommand \@sanitize@url [0]{\catcode `\\12\catcode `\$12\catcode `\&12\catcode `\#12\catcode `\^12\catcode `\_12\catcode `\%12\relax}%
\providecommand \@@startlink[1]{}%
\providecommand \@@endlink[0]{}%
\providecommand \url  [0]{\begingroup\@sanitize@url \@url }%
\providecommand \@url [1]{\endgroup\@href {#1}{\urlprefix }}%
\providecommand \urlprefix  [0]{URL }%
\providecommand \Eprint [0]{\href }%
\providecommand \doibase [0]{https://doi.org/}%
\providecommand \selectlanguage [0]{\@gobble}%
\providecommand \bibinfo  [0]{\@secondoftwo}%
\providecommand \bibfield  [0]{\@secondoftwo}%
\providecommand \translation [1]{[#1]}%
\providecommand \BibitemOpen [0]{}%
\providecommand \bibitemStop [0]{}%
\providecommand \bibitemNoStop [0]{.\EOS\space}%
\providecommand \EOS [0]{\spacefactor3000\relax}%
\providecommand \BibitemShut  [1]{\csname bibitem#1\endcsname}%
\let\auto@bib@innerbib\@empty
\end{thebibliography}%
\putbib[cites]
\end{bibunit}


\clearpage
\onecolumngrid
\appendix
\setcounter{page}{1} 
\resetlinenumber[1] 
\renewcommand{\theequation}{A\arabic{equation}}
\setcounter{equation}{0}
\renewcommand{\thetable}{S\arabic{table}}
\setcounter{table}{0}
\renewcommand{\thefigure}{S\arabic{figure}}
\setcounter{figure}{0}
\renewcommand{\thesubsection}{\Alph{subsection}}
\setcounter{subsection}{0}
\renewcommand{\thesection}{Appendix \Alph{section}}
\setcounter{section}{0}

\setcounter{secnumdepth}{1}
\tableofcontents
\begin{bibunit}

\section{Code Availability}

Our basin probing code is available at: \url{https://github.com/GilpinLab/loopscape}

\section{Supplementary Methods}

All experiments probe publicly-released recurrent reasoning models for which the model's latent state and its decoded output can be extracted at every reasoning loop. In all cases the reasoning model and problem instance jointly define an autonomous discrete-time dynamical system acting on the latent state, and we decoded the latent state after each loop as an intermediate representation of the solution.

\noindent\textbf{Equilibrium Reasoners (EqR)}~\cite{huang2026equilibrium} iterate paired latents $(z_H, z_L)$ over a sequence of $97$ tokens (for Sudoku, for which the hidden dimension is $512$) or $916$ tokens (for mazes, for which the hidden dimension is $128$). We use the authors' released checkpoints trained on Sudoku-Extreme and Maze-Unique. Because EqR is trained with stochastic inference, we disable its injected noise and fix the loop budget at $24$ iterations, and so the inference dynamics correspond to a deterministic map of the initial latent state. 

\noindent\textbf{Fixed-Point Reasoning Models (FPRM)}~\cite{movahedi2026fixedpointreasonersstableadaptive} iterate a single latent state toward a fixed point, halting once the relative residual falls below a fixed threshold $\tau$. We use the final released checkpoints and evaluation setting. For Sudoku, we set $\tau = 0.02$ with a $1000$-iteration maximum budget. For mazes we set $\tau = 0.1$ with a $100$-iteration cap. 

\noindent\textbf{Parcae}~\cite{prairie2026parcaescalinglawsstable} is a 140M parameter looped language model that we finetuned on the Countdown arithmetic task. We treat early decoding of text describing the solution as the puzzle output. The original model released by the authors was pretrained with an effective timescale of $\Delta t = 1.0$ associated with each latent iteration. We fine-tune this model with a smaller timestep of $\Delta t = 0.3$.

\noindent\textbf{Basin maps of initial-condition space.} For each model and problem instance, we hold the model, prompt, and loop schedule fixed and vary only the initial latent state along a random two-dimensional slice of the full latent space. We sample two orthonormal directions from the flattened latent space (e.g.\ $97 \times 512$ dimensions for Sudoku), using QR decomposition of a random Gaussian matrix initialized in the space.
We initialize the reasoning model at $z(a,b) = z_0 + a u + b v$ on a uniform grid
$(a,b) \in [-h,h]^2$ where $h$ is a half-width chosen for each model, and we use a resolution of $200 \times 200$ when computing population-level statistics.
The base point $z_0$ is a fixed random Gaussian sample in the space, and we scale $(a, b)$ based on the scale of each reasoning model's typical latent dynamics. For FPRM mazes, the random slice is confined to the fifteen "register" tokens that receive no input injection.

For every initial condition, we record the decoded output (solution state) at each loop. We define the convergence (settling) time as the number of loops until the decoded output of a given trajectory stops changing. Initial conditions still changing at the loop cap are marked as errors. The resulting settling time field over the two-dimensional slice constitutes the basin map.

\textbf{Task instances and difficulty.} Sudoku instances are drawn uniformly from the Sudoku-Extreme test set, using the benchmark's annotated per-puzzle difficulty ratings, which are based on backtracking encountered by a classical solver \cite{wang2025hierarchicalreasoningmodel}. FPRM maze instances are drawn from the Maze-Unique and Maze-Hard $30\times30$ test splits, with difficulty annotated by shortest-path length. For EqR maze experiments we use a set of $300$ generated perfect mazes, which correspond to random depth-first-search spanning trees on a $30\times30$ board. The wall fraction of these generated mazes is $\approx 0.57$, and the solution path lengths vary between $101$--$169$ steps. A maze with a unique shortest path can optionally be degenerated by opening $k$ pairwise-independent corner walls, each exactly doubling the number of optimal routes ($2^k$ shortest paths). Each sweep covers $100$--$300$ instances per random seed, with independent slice orientations across seeds.

\textbf{Quantifying basin fractality.} For each sampled two-dimensional slice of settling times we compute the basin entropy~\cite{daza2016basin}: we partition the slice into non-overlapping boxes, and compute the Shannon entropy $-\sum_i p_i \ln p_i$ of the settling times within each box. The basin entropy $S_b$ is the mean of this quantity over all boxes. A related quantity, the boundary entropy $S_{bb}$, averages only boxes containing more than one value. We also estimate the uncertainty exponent~\cite{mcdonald1985fractal} from the fraction $f(\varepsilon)$ of pixel pairs at separation $\varepsilon = 1, \ldots, 5$ pixels whose settling times differ. The basin-boundary dimension is $2 - \alpha$. Sampled slices are excluded from population statistics when fewer than $90\%$ of initial conditions reach the model's final solution (corresponding to cases in which the model had multistable or inconsistent solves). We also exclude slices where more than $1\%$ of pixels reach the maximum number of loops (corresponding to cases where simulating the dynamics to convergence is not computationally feasible). This produces at least $300$ valid slices for problems within each pair of reasoning models and tasks, from which we calculate correlations between the mean settling time across a slice, and $S_b$, $S_{bb}$, and $\alpha$.

\textbf{Fast Lyapunov Indicator.} We localize the boundaries separating distinct routes to equilibrium by calculating the fast Lyapunov indicator~\cite{froeschle1997fast} directly on the solution trajectories associated with two-dimensional slices of initial conditions. At each loop, we compare every pixel's decoded state under the dynamics at that time with the decoded state of the pixel's initial nearest neighbors. The maximum Euclidean norm of this time-dependent local deviation estimates $\lambda_F$. Large $\lambda_F$ therefore marks initial conditions with trajectories that separate maximally from their immediate neighbors before reconverging at the solution, thus corresponding to points straddling a saddle-like boundary between solution routes.

\textbf{Trajectory geometry and decoding near saddles.} To study the geometry of trajectories, we analyze the full latent dynamics of a set of initial conditions (and not just the dynamics in the solution space, as in previous analyses). At each iteration, we compute the residual between the latent timepoint and the final latent solution. For a given puzzle, we perform principal components analysis on all latent trajectories, pooled across both initial conditions and timepoints.

\section{Replicate basin experiments with alternative reasoning models}
\label{app:replicates}

\begin{figure*}[!ht]
{
\centering
\includegraphics[width=\linewidth]{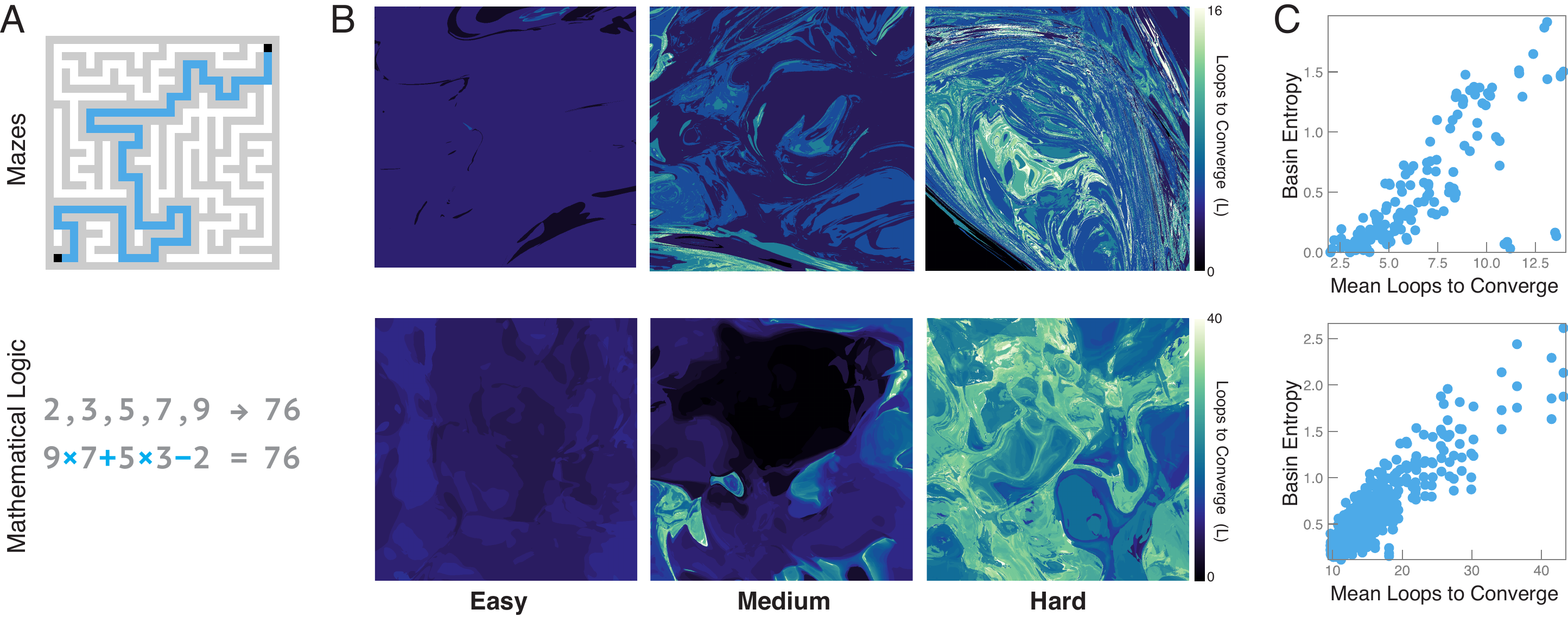}
\caption{
\textbf{Hard problems produce fractal basins across alternative models.}
(A) Examples of Maze-Hard and Countdown reasoning tasks, each of which we probe using a different architecture (Equilibrium Reasoners and a looped transformer) \cite{wang2025hierarchicalreasoningmodel,huang2026equilibrium}
(B) Example basins for varying problem difficulties across the two tasks and architectures.
(C) Scaling of basin entropy with the number of reasoning iterations required to converge, each across randomly-sampled task instances of varying underlying difficulty.
}
\label{fig:extra}
}
\end{figure*}

In order to rule out fractal basins only arising from certain task-model pairings, we replicate our findings using alternative architectures to solve the Maze and Countdown tasks (Fig. \ref{fig:extra}). 
We consider an Equilibrium Reasoning model checkpoint trained by the original authors to solve maze tasks, and we train an encoder-only looped transformer to solve Countdown \cite{huang2026equilibrium}. 
We find that fractal basins emerge on difficult problems, and we observe a relationship between fractality and task difficulty.
As we observed with other models, the qualitative appearance and structure of the basins change, likely because they are a property of the particular optimization algorithm that each model uses to solve a task.

\section{Metrics for basin complexity}
\label{app:metrics}

\begin{figure*}[ht]
{
\centering
\includegraphics[width=\linewidth]{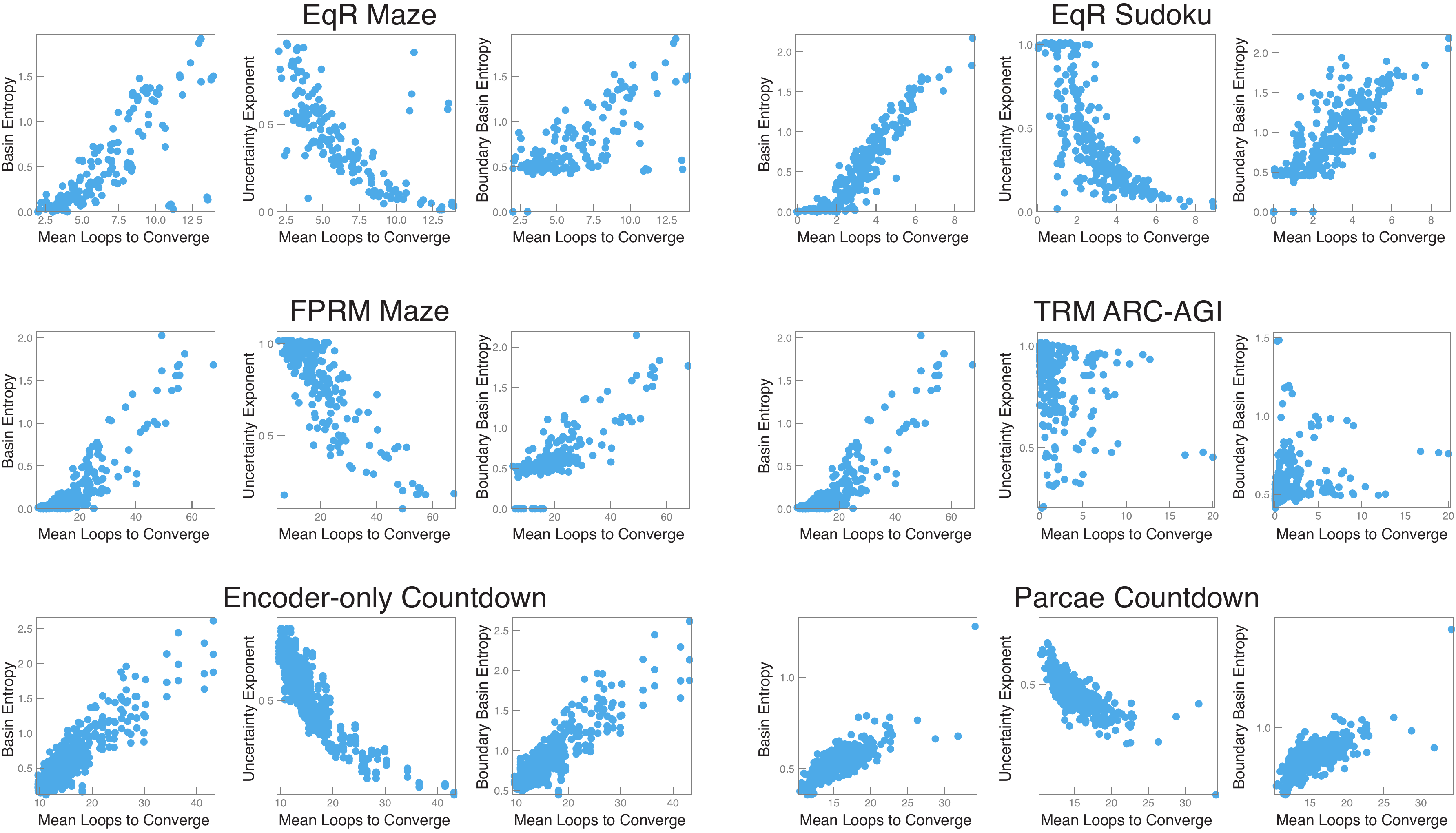}
\caption{
\textbf{Alternative metrics of basin complexity.}
We repeat the basin quantification procedure using three alternative metrics. Each point corresponds to metrics calculated over a random two-dimensional slice through a random problem instance.
}
\label{sfig:metrics}
}
\end{figure*}

Across all tasks and architectures, we compute three measures of basin complexity (Fig. \ref{sfig:metrics}). 
The basin entropy measures the average variation in the convergence time within finite-resolution neighborhoods of different initial conditions, while the boundary basin entropy computes this quantity only over a subset of initial conditions straddling multiple basins \cite{daza2016basin}.
We also compute the uncertainty exponent, which measures how the fraction $f(\varepsilon)$ of initial conditions whose convergence time changes under a perturbation of size $\varepsilon$ scales as $f(\varepsilon)\sim\varepsilon^\alpha$ \cite{grebogi1983final}. Equivalently, $\alpha=D-d_b$, where $D$ is the dimension of the sampled initial-condition space ($D=2$ in our experiments) and $d_b$ is the boundary's box-counting dimension. Due to this relationship, we broadly observe that the different metrics are all strongly correlated across different puzzle instances and tasks.

\section{Measuring fractality across scales}
\label{sec:zoom}

We show the fractal structure of different reasoning models and tasks by repeatedly zooming into regions of the space of initial conditions. We show plots of the convergence time and the fast Lyapunov indicator at each scale, across different model architectures \cite{froeschle1997fast}. In order to quantify the scale dependence of fractality, for each model and task we measure the uncertainty exponent \cite{grebogi1983final}, which measures how variation in outcomes within a fixed-radius-$\epsilon$ ball of initial conditions varies with $\epsilon$. For true fractals, uncertainty and radius should exhibit a fixed power-law relationship, supporting the standard view of fractals as scale-free objects \cite{mandelbrot1967long}.

We find that some reasoning models produce true fractals, and thus have similar degrees of qualitative complexity when one zooms in. Other reasoning models, however, produce recently-characterized "slim fractals," which correspond to the fractal dimension decreasing with the level of resolution \cite{chen2017slim}. Slim fractals are scale-dependent, and the true basin boundary has codimension $1$. They typically arise in physical systems with dissipation, where the saddle-like set that normally scatters trajectories (and causes transient chaos) decays over time \cite{motter2013doubly,omel2022focusing}. For example, a double pendulum has a sustained chaotic attractor in the absence of friction. However, when friction is introduced, energy gradually leaves the system, and the only long-term attractor is a fixed point (in which the pendulum is at rest) \cite{karolyi2021new}. At any given instant (and energy level) the pendulum has a saddle-like set, but this set decays over time. This general phenomenon, doubly-transient chaos, differs from singly-transient chaos (such as scattering fields in the three-body problem) in which the saddle-like set is constant, and transience arises purely from trajectories approaching non-chaotic steady states at long times \cite{stone2019statistical,wiggins1994normally}.

Reasoning models are not typically designed to directly map onto physical systems, and so we generally expect that their dynamics can produce types of basins and transient chaos beyond singly- or doubly-transient. However, many properties built into these models, such as a tendency to converge to fixed answers, likely represent a form of dissipation, and so we expect that most reasoning models will exhibit some form of doubly transient chaos.


\begin{figure}[h]
    \centering
    \includegraphics[width=1.0\linewidth]{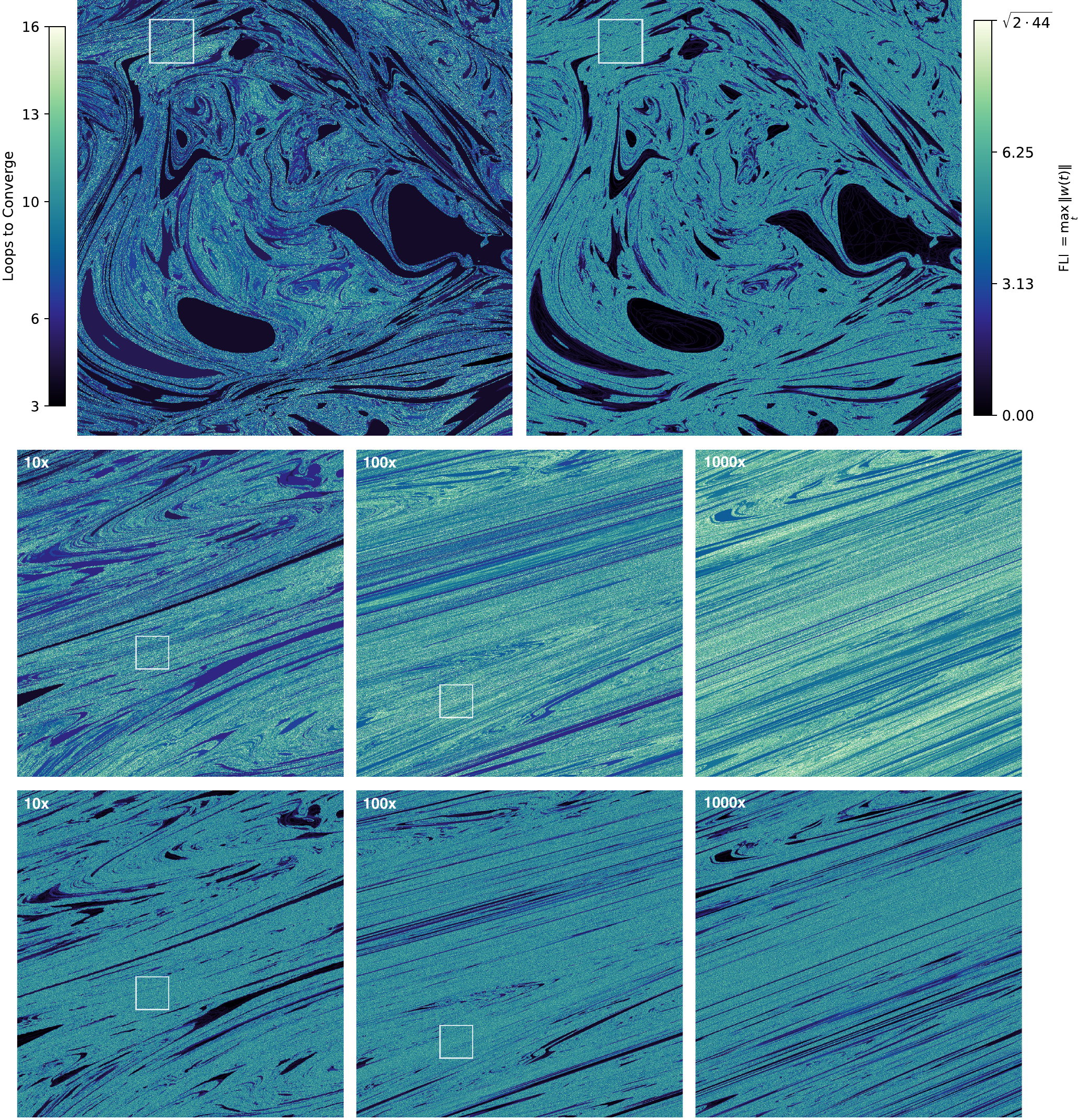}
    \caption{Progressive magnifications of a fractal produced by the Equilibrium Reasoning model on a difficult Sudoku task. Top Left: settle-time field (loops to converge); Top Right: fast Lyapunov indicator (FLI) on the settle-time field. Middle Row: zoom sequence on the settle-time field. Bottom Row: zoom sequence on the FLI field.}
    \label{fig:eqr-sudoku_prompt-A}
\end{figure}


\begin{figure}
    \centering
    \includegraphics[width=0.7\linewidth]{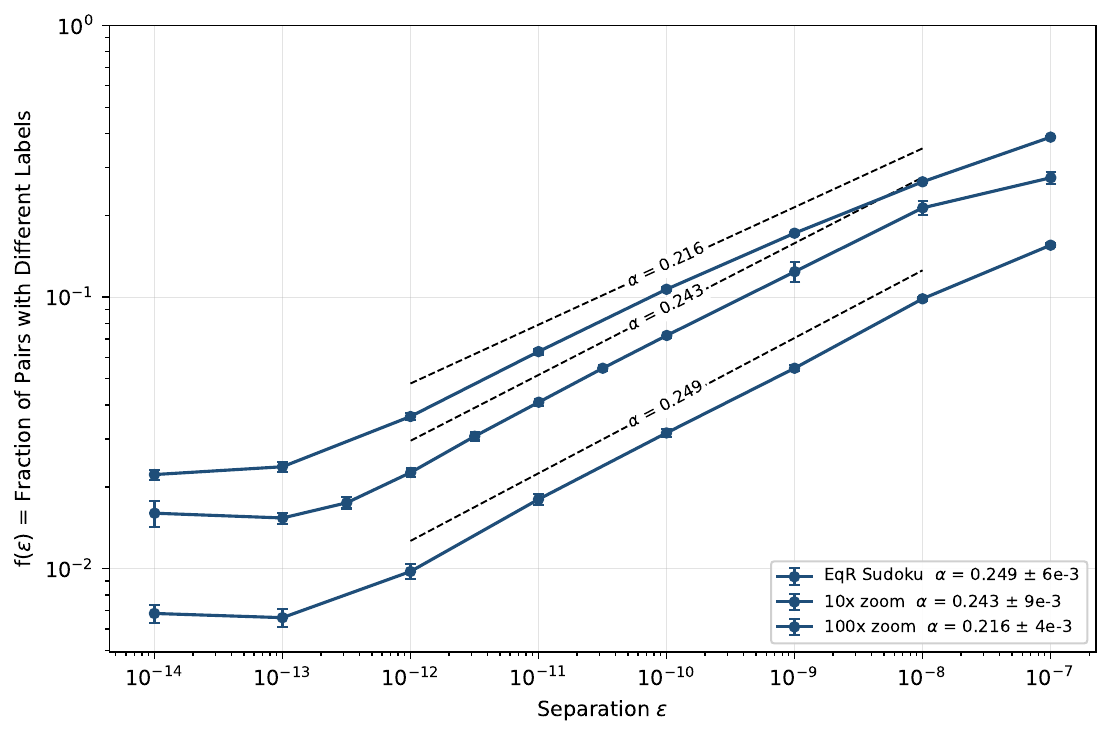}
    \caption{Scaling of the uncertainty exponent for the fractal in Fig. \ref{fig:eqr-sudoku_prompt-A}.}
    \label{fig:eqr_sudoku_uncertainty_exponent_scaling}
\end{figure}

\begin{figure}[h]
    \centering
    \includegraphics[width=1.0\linewidth]{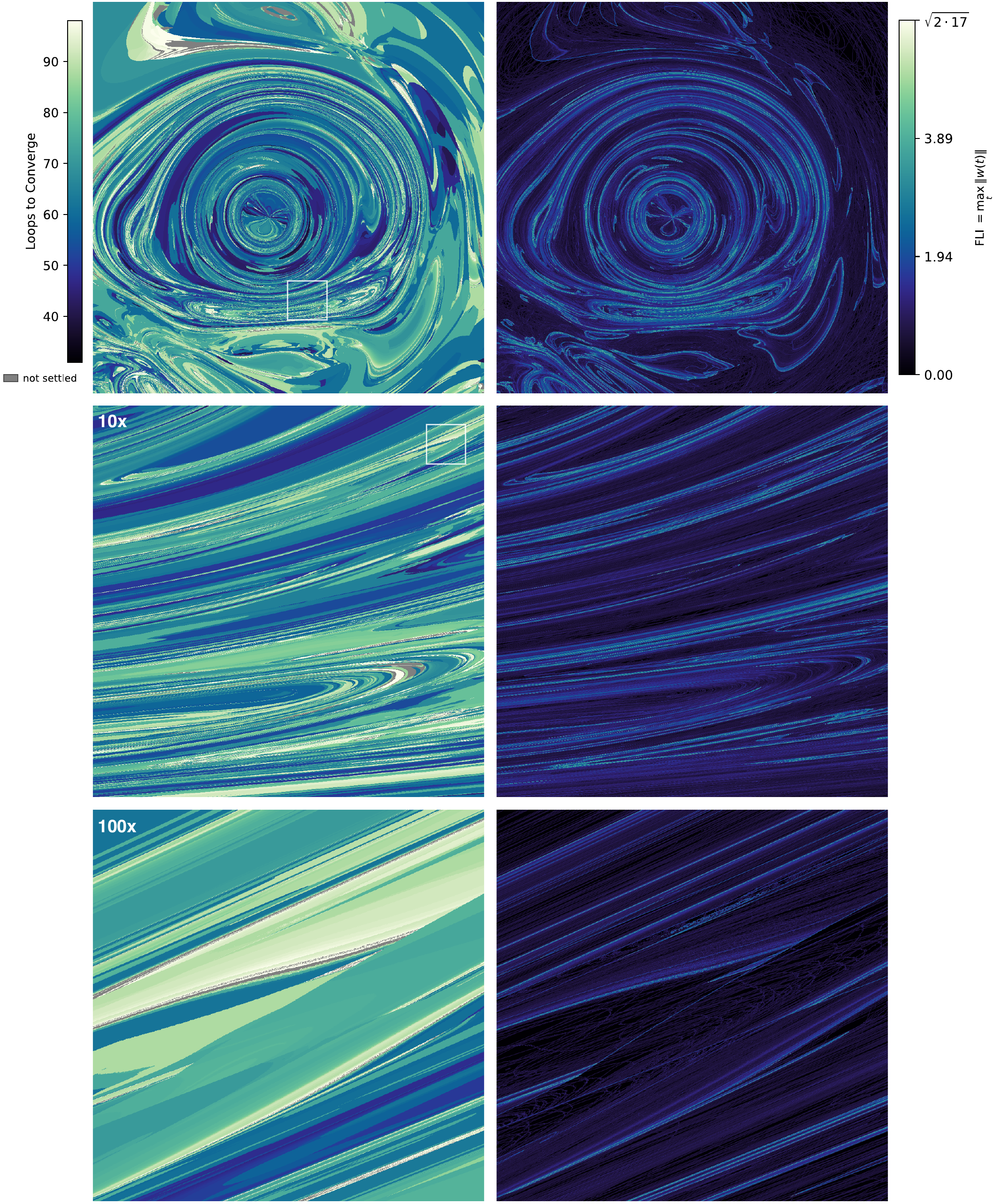}
    \caption{Progressive magnification of a fractal produced by FPRM on the Maze task.}
    \label{fig:fprm_maze_zoomseq}
\end{figure}

\begin{figure}
    \centering
    \includegraphics[width=0.6\linewidth]{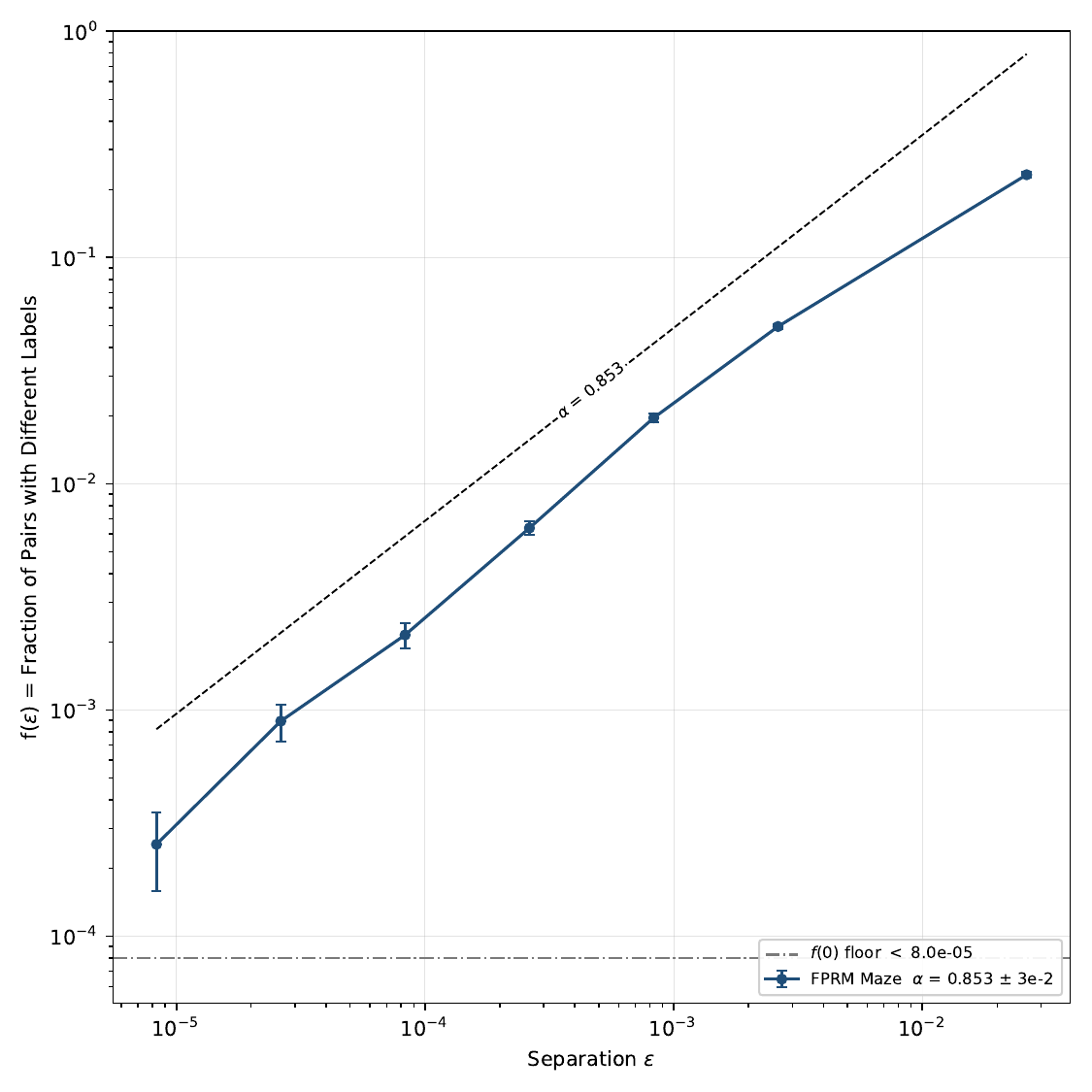}
    \caption{Scaling of the uncertainty exponent for the fractal in Fig. \ref{fig:fprm_maze_zoomseq}.}
    \label{fig:fprm_maze_uncertainty_exponent_scaling}
\end{figure}


\begin{figure}
    \centering
    \includegraphics[width=1.0\linewidth]{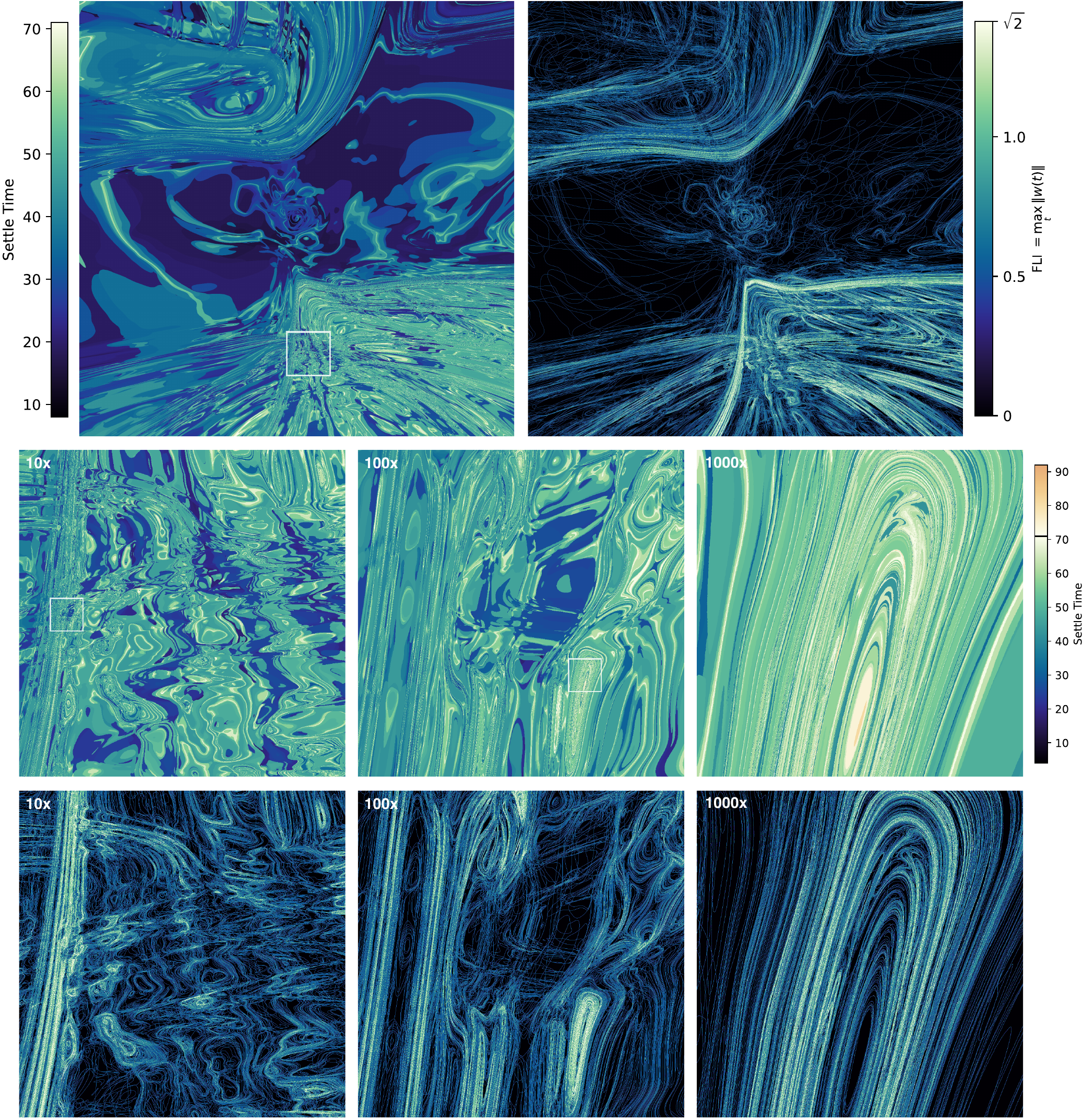}
    \caption{Progressive magnification of a fractal produced by the fine-tuned Parcae model on the Countdown mathematical logic task.}
    \label{fig:ood139_s1}
\end{figure}


\begin{figure}
    \centering
    \includegraphics[width=0.85\linewidth]{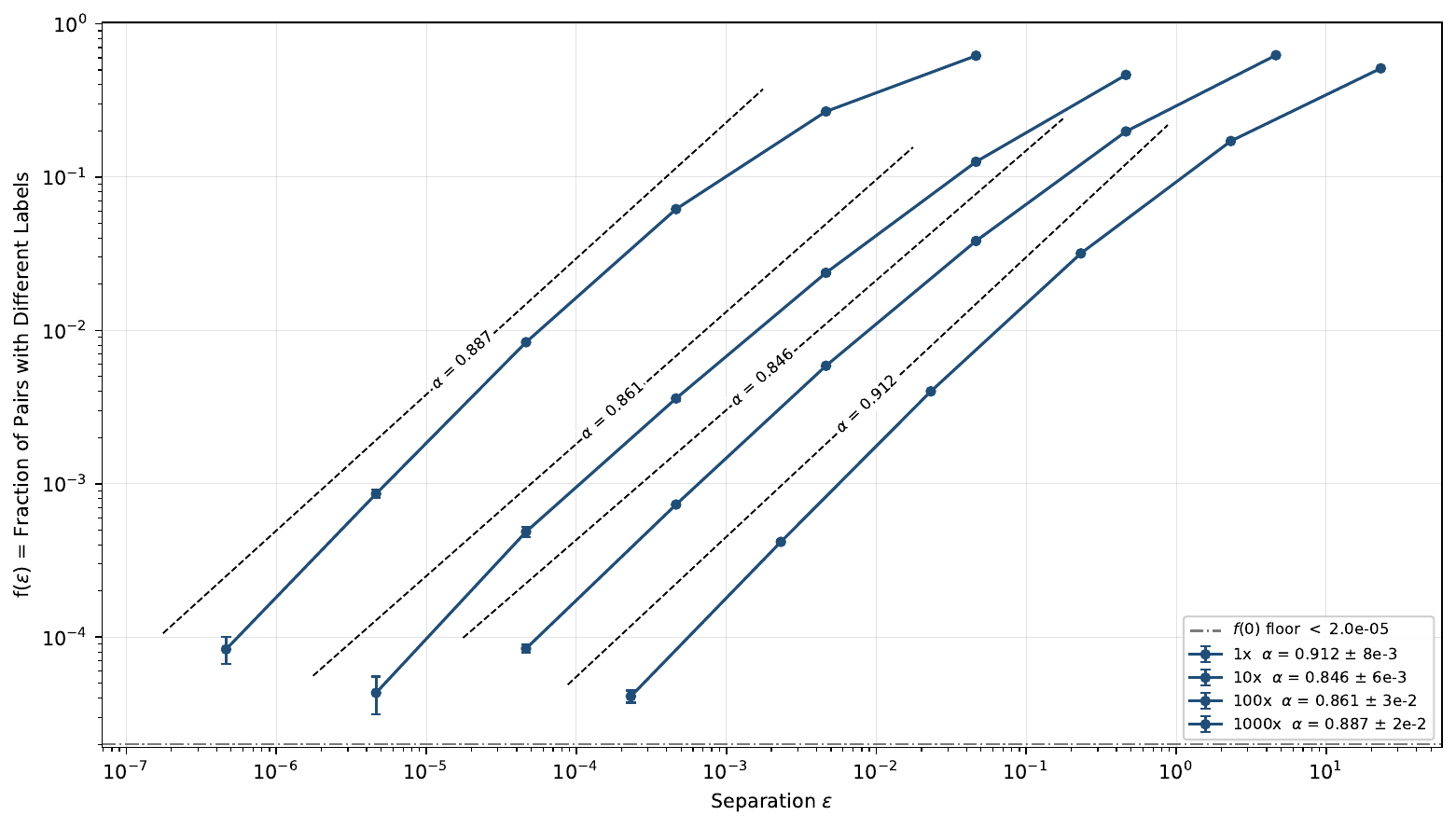}
    \caption{Scaling of the uncertainty exponent for the fractal in Fig. \ref{fig:ood139_s1}.}
    \label{fig:parcae_countdown_uncertainty_exponent_scaling}
\end{figure}

\clearpage
\putbib[cites]
\end{bibunit}
\end{document}